\documentclass{article}

\usepackage{microtype}
\usepackage{graphicx}
\usepackage{subcaption}
\usepackage{multirow}
\usepackage{booktabs} 
\usepackage[table,xcdraw]{xcolor}

\usepackage{hyperref}

\usepackage[preprint]{icml2026}

\usepackage{amsmath}
\usepackage{amssymb}
\usepackage{mathtools}
\usepackage{amsthm}

\usepackage[capitalize,noabbrev]{cleveref}

\theoremstyle{plain}

\theoremstyle{definition}

\theoremstyle{remark}

\usepackage[textsize=tiny]{todonotes}

\icmltitlerunning{Submission and Formatting Instructions for ICML 2026}

\begin{document}

\twocolumn[
  \icmltitle{Less Experts, Faster Decoding: Cost-Aware 
\texorpdfstring{\\}{ }Speculative Decoding for Mixture-of-Experts}



  \icmlsetsymbol{equal}{*}
  \icmlsetsymbol{intern}{$\ddagger$}

  \begin{icmlauthorlist}
    \icmlauthor{Jincheng Xie}{equal,thu,intern}
    \icmlauthor{Runheng Liu}{equal,bit,intern}
    \icmlauthor{Heyan Huang}{bit}
    \icmlauthor{Yawen Ling}{jd}
    \icmlauthor{Hanbin Dai}{jd}
    \icmlauthor{Yu Zheng}{xnju,xdu}
    \icmlauthor{Wen Hu}{jd}
    
  \end{icmlauthorlist}

  \icmlaffiliation{thu}{Department of Mathematical Sciences, Tsinghua University, Beijing, China}
  \icmlaffiliation{bit}{School of Computer Science and Technology, Beijing Institute of Technology, Beijing, China}
  \icmlaffiliation{xnju}{School of Computing and Artificial Intelligence, Southwest Jiaotong University, Chengdu, Sichuan, China}
  \icmlaffiliation{xdu}{School of Cyber Engineering, Xidian University, Xi'an, Shaanxi, China}
  \icmlaffiliation{jd}{JDT AI Infra}
  
  \icmlcorrespondingauthor{Wen Hu}{huwen.31@jd.com}

  \icmlkeywords{Machine Learning, ICML}

  \vskip 0.3in
]



\printAffiliationsAndNotice{
\icmlEqualContribution
\textsuperscript{$\ddagger$}Work done while interning at JDT AI Infra.
}

\begin{abstract}
Sparse Mixture-of-Experts (MoE) models have become an important approach for scaling Large Language Models (LLMs), but their inference efficiency depends strongly on expert activation patterns. Speculative decoding (SD) accelerates autoregressive generation by verifying multiple draft tokens in parallel, yet existing draft selection strategies primarily optimize acceptance likelihood. In large-scale MoE models, however, selecting draft tokens also determines the union of experts activated during verification. We observe that confidence-driven SD can introduce \textit{expert scattering}: high-probability draft tokens may route to disjoint experts, increasing expert-weight memory traffic and reducing the speedup from speculation. Motivated by this observation, we revisit draft-tree selection under the non-uniform memory-cost structure of MoE inference. We propose \textsc{EcoSpec}, a cost-aware speculative decoding framework that incorporates predicted marginal expert activation cost into draft selection. With a lightweight expert predictor and a dynamic expert buffer, \textsc{EcoSpec} favors draft paths that preserve high acceptance likelihood while reusing experts already covered by the current verification set, without modifying the target-model verification rule. We evaluate \textsc{EcoSpec} on three large-scale MoE models, including DeepSeek-V3.1 (671B), Qwen3-235B-A22B, and GPT-OSS-120B, across reasoning, coding, question-answering, and dialogue benchmarks. \textsc{EcoSpec} consistently reduces active expert footprints and improves end-to-end decoding speed, achieving up to $1.62\times$ speedup. These results show that accounting for expert activation cost is important for efficient speculative decoding in large-scale MoE models.
\end{abstract}

\section{Introduction}
\label{sec:introduction}

\begin{figure*}[t]
    \centering
    \begin{subfigure}[b]{0.32\textwidth}
        \centering
        \includegraphics[width=\textwidth]{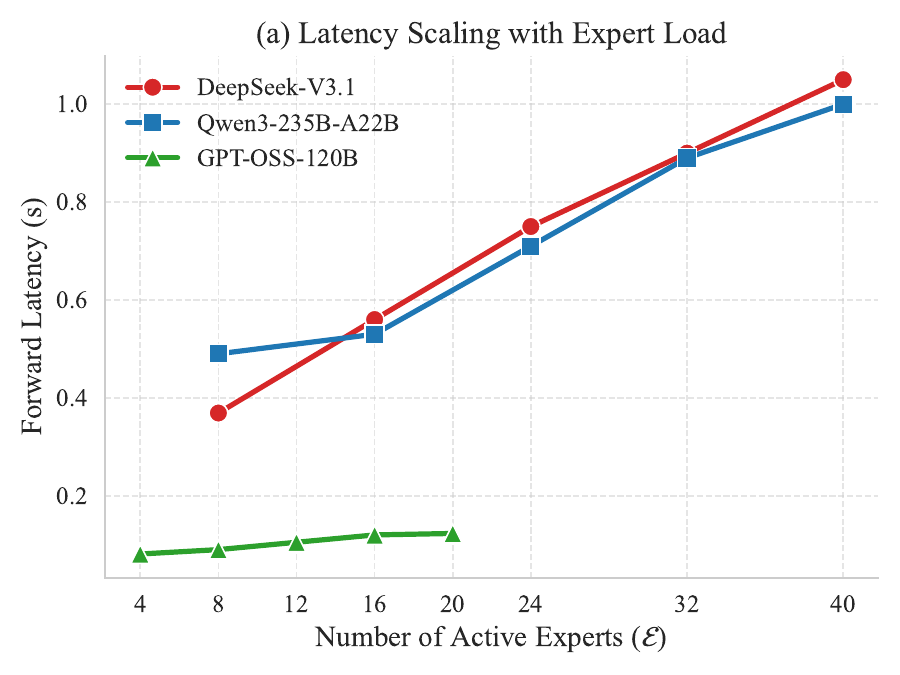}
        \caption{\textbf{Latency vs. Expert Load}}
        \label{fig:latency}
    \end{subfigure}
    \hfill
    \begin{subfigure}[b]{0.32\textwidth}
        \centering
        \includegraphics[width=\textwidth]{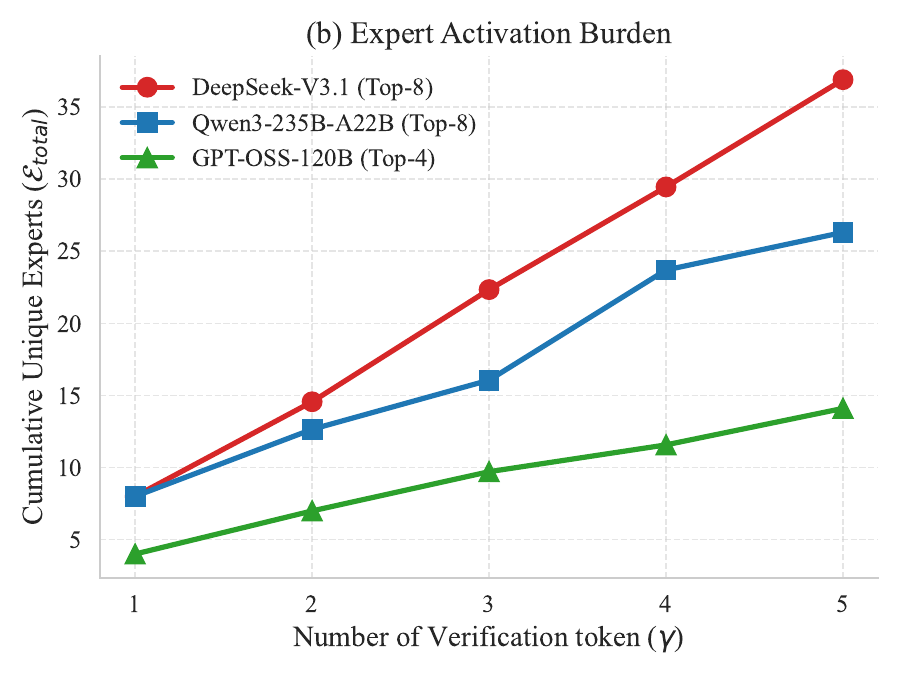}
        \caption{\textbf{Expert Activation Burden}}
        \label{fig:burden}
    \end{subfigure}
    \hfill
    \begin{subfigure}[b]{0.32\textwidth}
        \centering
        \includegraphics[width=\textwidth]{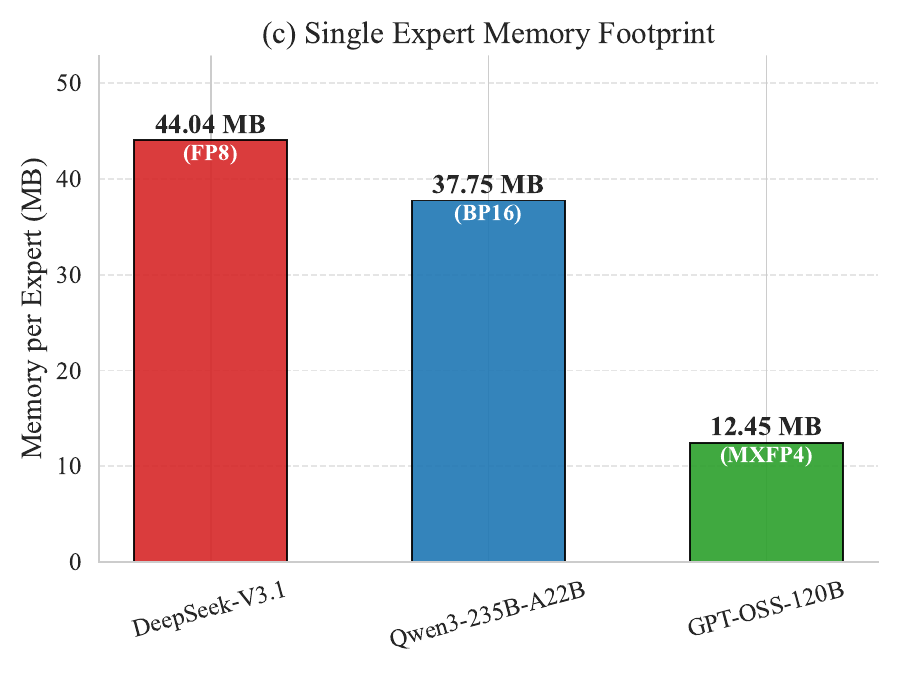}
        \caption{\textbf{Single Expert Cost}}
        \label{fig:memory}
    \end{subfigure}
    
    \caption{\textbf{The Bandwidth Bottleneck in MoE Speculative Decoding.} 
    (a) Verification latency scales linearly with the number of active experts ($\mathcal{E}$), creating a strict latency penalty for retrieving extra experts.
    (b) Top-K means K activated experts in each layer. As the verification budget $\gamma$ increases, standard baselines (e.g., Eagle) rapidly activate disjoint sets of experts, quickly saturating the memory bandwidth.
    (c) The physical memory footprint of a \textit{single} expert is substantial. For DeepSeek-V3.1 (FP8), loading just one expert consumes \textbf{44.04 MB} of HBM bandwidth, implying that every mispredicted expert path incurs a massive I/O overhead.}
    \label{fig:motivation}
\end{figure*}

Large Language Models (LLMs) have demonstrated remarkable capabilities across logic, coding, and creative tasks \cite{brown2020language, achiam2023gpt, grattafiori2024llama3herdmodels}, yet their deployment is increasingly constrained by high inference latency and serving cost \cite{deepspeed}. The standard autoregressive decoding process generates tokens sequentially, where each step requires a full pass through the target model \cite{vaswani2017attention}. As models scale to hundreds of billions of parameters, this serial dependency makes inference heavily memory-bound: the arithmetic intensity is low, and throughput is often limited by the bandwidth required to load model weights from High Bandwidth Memory (HBM) to on-chip compute units \cite{williams2009roofline, shazeer2019fast, dao2022flashattention}. Consequently, reducing decoding latency and memory traffic has become a critical priority for both real-time user experience and infrastructure efficiency.

To mitigate the serial decoding bottleneck, Speculative Decoding (SD) has been widely studied as an inference acceleration paradigm \cite{leviathan2023fast, chen2023accelerating, li2024eagle, li2024eagle2, li2025eagle3, cai2024medusa}. 
SD uses a low-cost draft mechanism to propose multiple candidate tokens, which are then verified in parallel by the target model. 
When the drafts are accepted, a single target-model forward pass can advance generation by multiple tokens, thereby amortizing parameter loading and improving hardware utilization. 
For dense Transformers, where the same weight matrices are reused across all verified positions, this parallel verification can substantially increase arithmetic intensity and reduce the effective cost per generated token \cite{miao2023specinfer}.

However, the dense-model assumption behind this amortization does not directly extend to sparse Mixture-of-Experts (MoE) architectures \cite{shazeer2017outrageously, fedus2022switch}, which are increasingly adopted in large-scale language models \cite{gpt-oss, deepseek2024v3, qwen3}. 
In MoE layers, the dense feed-forward module is replaced by a pool of experts, and a router assigns each token to a small top-$k$ subset of experts \cite{lepikhin2020gshard, du2022glam}. 
Therefore, parallel verification no longer reuses a single fixed set of feed-forward weights across all verified positions. Its memory cost depends on the union of experts activated by the verified tokens: when different candidate tokens route to disjoint experts, the verifier must fetch additional expert weight blocks from HBM. As a result, verification latency in MoE speculative decoding becomes highly sensitive to expert overlap and reuse, rather than being determined mainly by the number of verified tokens.

This MoE-specific cost structure exposes a mismatch between the acceptance-driven selection objective of existing SD methods and the memory cost of MoE verification. Many recent SD methods organize draft candidates as a tree and select a verification subset primarily according to confidence or acceptance likelihood. This criterion is effective for dense models, where the main objective is to maximize the number of accepted tokens per target-model pass. In MoE verification, however, a high-probability candidate can still activate experts that are disjoint from those used by other verified tokens. Adding such a candidate may therefore enlarge the per-step expert union, increase expert-weight memory traffic, and reduce cache reuse \cite{Huang2025MoESDUS, xue2024moe}. We refer to this expansion of the per-step expert footprint as \textit{expert scattering}. As shown in Fig.~\ref{fig:motivation}, verification latency increases with the active expert footprint, while confidence-driven selection can rapidly increase the number of unique experts touched within one verification step. Consequently, expanding the verification set may erode end-to-end speedup even when acceptance rates remain comparable. This observation motivates a cost-aware draft selection objective that balances acceptance likelihood with the marginal cost of introducing new experts.

To address this issue, we propose \textsc{EcoSpec}, a cost-aware speculative decoding framework for MoE models. \textsc{EcoSpec} operates at the draft-tree selection stage, where it selects candidates for parallel verification under an acceptance--cost trade-off. Instead of ranking draft tokens primarily by acceptance likelihood, \textsc{EcoSpec} also accounts for the marginal expert cost induced by each candidate. This allows the selection procedure to prefer draft paths that maintain high acceptance probability while reusing experts already covered by the current verification set. Importantly, \textsc{EcoSpec} does not modify the target-model verification rule and therefore preserves the lossless semantics of standard speculative decoding \cite{leviathan2023fast}. By aligning draft selection with the memory-cost structure of MoE inference, \textsc{EcoSpec} reduces unnecessary expert-weight traffic during verification. Empirically, \textsc{EcoSpec} achieves consistent speedups across multiple production-scale MoEs, including up to \textbf{1.62$\times$} on Qwen3-235B, \textbf{1.50$\times$} on GPT-OSS-120B, and \textbf{1.47$\times$} on DeepSeek-V3.1.

Our contributions are summarized as follows:
\begin{itemize}
    \item We identify and analyze \textit{expert scattering} in large-scale MoE speculative decoding: confidence-driven draft selection can expand the per-step union of activated experts, increasing expert-weight memory traffic during verification.
    \item We propose \textsc{EcoSpec}, a cost-aware speculative decoding framework that incorporates marginal expert activation cost into draft-tree selection. With a lightweight expert predictor and a dynamic expert buffer, \textsc{EcoSpec} favors draft paths that preserve high acceptance likelihood while improving expert reuse, without changing the standard lossless verification procedure.
    \item We evaluate \textsc{EcoSpec} on three large-scale MoE models---DeepSeek-V3.1 (671B), Qwen3-235B-A22B, and GPT-OSS-120B---across diverse reasoning, coding, and dialogue benchmarks. \textsc{EcoSpec} consistently reduces activated experts and improves end-to-end decoding speed, achieving up to $1.62\times$ speedup over existing SD baselines.
\end{itemize}

\begin{figure*}[t]
    \centering
    \includegraphics[width=1.0\textwidth]{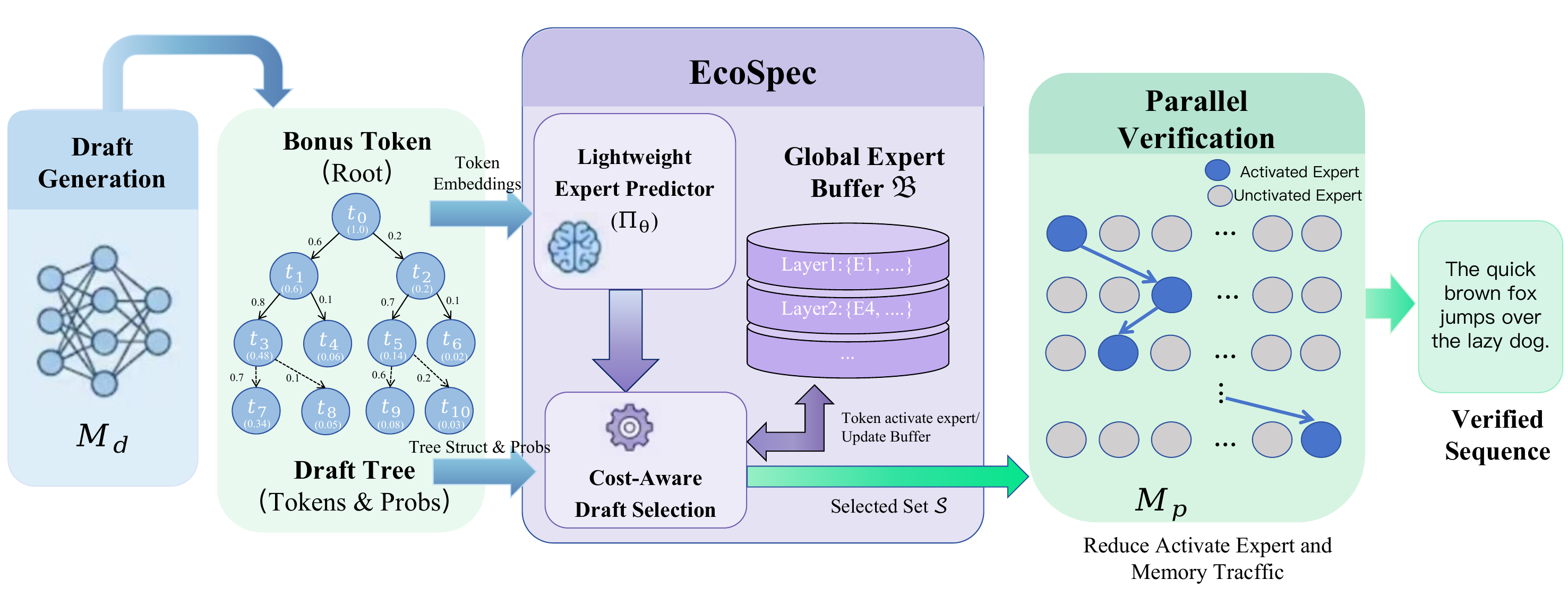}
    
    \caption{\textbf{Overview of the \textsc{EcoSpec} framework.}
    The process begins with draft generation, where a draft model $M_d$ produces a tree of candidate tokens with associated probabilities.
    The \textsc{EcoSpec} module then selects a set of draft tokens $\mathcal{S}$ for verification.
    It employs a lightweight expert predictor $\Pi_\theta$ to estimate expert activations, maintains a global expert buffer $\mathcal{B}$ to track experts already covered by the selected tokens, and uses cost-aware draft selection to balance acceptance likelihood with expert activation cost.
    Finally, the target MoE model $M_p$ performs parallel verification on the selected set $\mathcal{S}$.
    }
    \label{fig:framework}
\end{figure*}

\section{Related Work}
\label{sec:related work}

\paragraph{Speculative Decoding and Multi-Token Prediction.}
Speculative decoding (SD) accelerates LLM inference by using a lower-cost draft mechanism to propose multiple tokens, which are then verified in parallel by the target model \cite{leviathan2023fast, chen2023accelerating}. Early approaches commonly use a separate draft model to generate candidate continuations \cite{miao2023specinfer}. Recent methods further improve draft efficiency and acceptance by adopting tree-structured drafting or auxiliary prediction heads, including Medusa and the EAGLE series \cite{cai2024medusa, li2024eagle, li2024eagle2, li2025eagle3}. In parallel, Multi-Token Prediction (MTP) introduces an auxiliary training objective that enables in-model token predictors to serve as draft heads for speculative verification \cite{deepseek2024v3, gloeckle2024better}. Despite architectural differences in how drafts are produced, these approaches typically prioritize draft tokens based on confidence (acceptance likelihood) and do not explicitly account for the hardware cost of activating additional experts in MoE verification. As a result, when applied to MoE models, high-confidence drafts may still trigger a rapidly expanding union of activated experts, corresponding to our Expert Scattering phenomenon and the associated memory inefficiency. EcoSpec makes this explicit by optimizing a cost-aware objective that balances acceptance likelihood against the incremental cost of activating new experts.

\paragraph{Efficient Inference for Mixture-of-Experts.}
Mixture-of-Experts (MoE) models increase model capacity while reducing per-token FLOPs, but their inference efficiency is often limited by expert-weight memory traffic and token--expert dispatch overheads \cite{shazeer2017outrageously, xue2024survey}.
Prior work improves MoE execution under a given routing pattern through expert caching and prefetching \cite{xue2024moeinfinity, huang2024expertbuffering}, optimized token dispatch and fused MoE kernels \cite{gale2023megablocks}, and routing or load-balancing strategies that reduce uneven expert utilization \cite{fedus2022switch, deepseek2024v3}.
Recent systems further study MoE inference in the speculative-decoding setting.
SP-MoE and MoE-SpeQ use speculative lookahead to support expert prefetching, offloading, and execution scheduling, aiming to hide or reduce expert-movement overhead during MoE serving \cite{chen2025spmoespeculativedecodingprefetching, wang2025moespeqspeculativequantizeddecoding}.
MoE-Spec instead reduces verification-time expert cost by imposing an expert budget and selecting only a subset of experts to load during speculative verification \cite{mcdanel2026moespecexpertbudgetingefficient}.
These methods either optimize runtime execution under predicted or routed expert demand, or change the verification-time expert budget.
\textsc{EcoSpec} addresses a different stage of the decoding pipeline: it preserves the standard target-model verifier and lossless speculative decoding semantics, and only changes which draft-tree nodes are selected before verification.
Therefore, \textsc{EcoSpec} is complementary to MoE runtime optimizations, since expert caching, prefetching, offloading, or optimized dispatch can still be applied after \textsc{EcoSpec} reduces the expert working set induced by the selected verification nodes.

\begin{figure*}[t]
    \centering
    \includegraphics[width=1.0\textwidth]{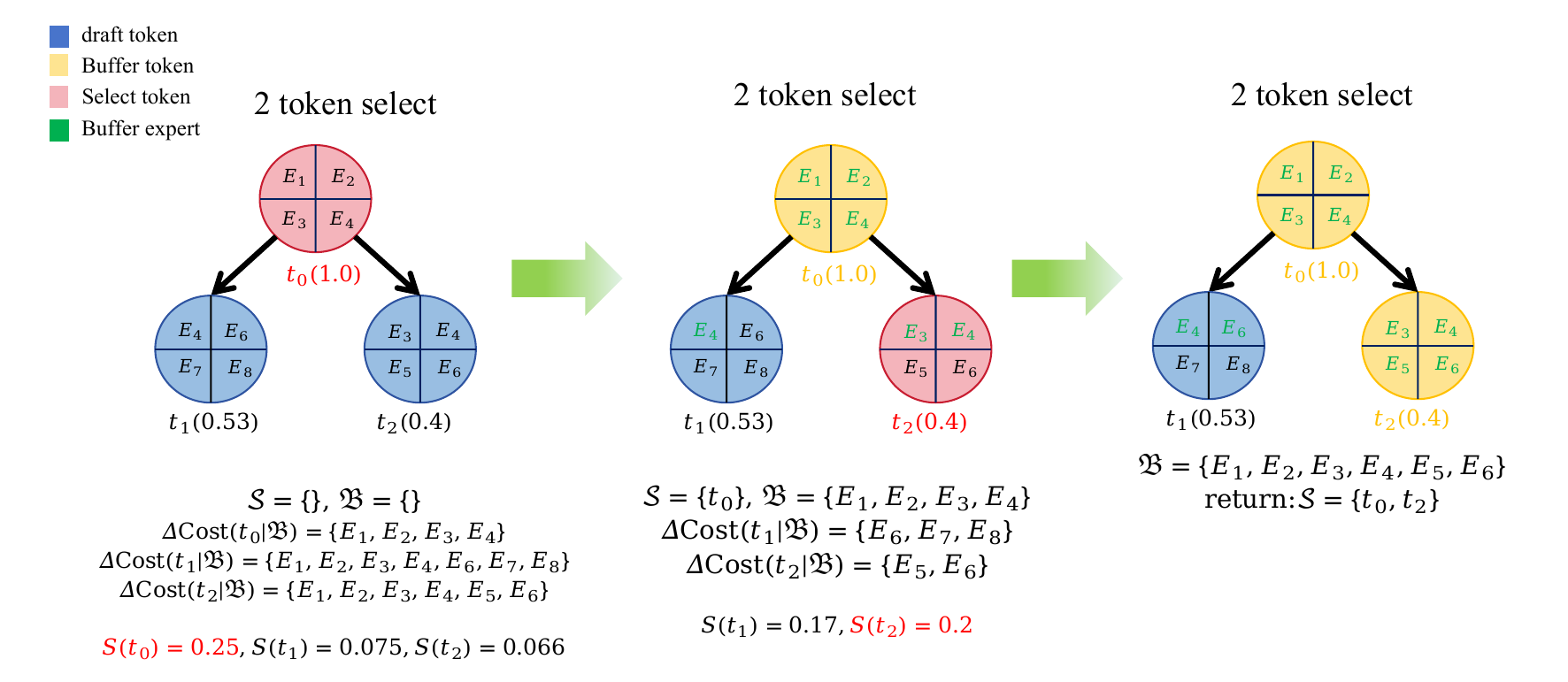}
    \caption{Illustration of Pre-verification Cost-Aware Subset Construction.
    We select $\gamma=2$ draft tokens before target-model verification. 
    Left (Step 1): the root token $t_0$ is selected according to the acceptance--cost score, and its predicted expert footprint $\{E_1,E_2,E_3,E_4\}$ is added to the expert buffer $\mathcal{B}$ for subsequent scoring. 
    Middle (Step 2): the algorithm re-evaluates the remaining candidates using the updated buffer. Although $t_1$ has higher cumulative draft probability ($P=0.53$), it introduces three new predicted experts ($\Delta \mathrm{Cost}=3$). In contrast, $t_2$ reuses experts already covered by $\mathcal{B}$ and introduces only two new predicted experts ($\Delta \mathrm{Cost}=2$), so it obtains a higher score ($S=0.20>0.17$) and is selected. 
    Right (Final): the final selected set is $\mathcal{S}=\{t_0,t_2\}$, illustrating how marginal expert cost can change the ranking induced by cumulative draft probability.}
    \label{fig:select}
\end{figure*}

\section{MoE Speculative Decode Bottleneck}
\label{sec:motivation}

Speculative Decoding (SD) accelerates autoregressive generation by verifying multiple draft tokens in a single target-model forward pass. For dense Transformers, the memory cost of verifying $\gamma$ tokens is close to that of verifying a single token, because all verified positions reuse the same dense weight matrices within the verification batch. Thus, increasing the verification budget mainly improves the amortization of target-model parameter loading and can reduce the effective cost per generated token.

This premise becomes less reliable for sparse MoE models. For a MoE layer $\ell$, let $\mathcal{S}_{\ell}(x_t)$ denote the set of experts activated by the $t$-th verified token. Verifying a draft sequence $x_{1:\gamma}$ requires accessing the union of experts activated by all verified tokens at that layer:
\begin{equation}
    \mathcal{E}^{\ell}_{\mathrm{verify}} 
    = \bigcup_{t=1}^{\gamma} \mathcal{S}_{\ell}(x_t).
\end{equation}
The verification footprint is therefore determined by the size of this union, aggregated across MoE layers, rather than by the number of verified tokens alone. If selected draft tokens route to overlapping experts, verification can reuse expert weights. If they route to disjoint experts, the verifier must fetch additional expert weight blocks from HBM, increasing memory traffic within the same speculative step.

This creates an expert-scattering effect in MoE speculative decoding. Standard SD selection strategies are typically designed to maximize acceptance likelihood and are not aware of expert locality. As a result, adding more high-confidence draft tokens can enlarge the per-step expert union even when those tokens are likely to be accepted. In practice, $|\mathcal{E}^{\ell}_{\mathrm{verify}}|$ can grow rapidly with the verification budget $\gamma$, reducing expert reuse and creating an additional memory-traffic bottleneck.

We profile DeepSeek-V3.1 (671B), Qwen3-235B-A22B, and GPT-OSS-120B on GSM8K \cite{gsm8k} under identical execution settings. Fig.~\ref{fig:motivation}(a) shows that verification latency increases with the number of unique experts activated per step, indicating that fetching additional expert weights is a major source of verification cost. Fig.~\ref{fig:motivation}(b) shows that standard SD selection can rapidly increase the cumulative number of unique activated experts as more tokens are selected for verification. This means that a larger verification set provides more opportunities for token acceptance, but can also make each verification pass more expensive in MoE models.

The cost of this effect is substantial at large scale. As shown in Fig.~\ref{fig:motivation}(c), a single expert can occupy tens of megabytes of memory, so even a small increase in the per-layer expert union can translate into large HBM traffic. For DeepSeek-V3.1, reducing the expected expert footprint by only $0.2$ experts per layer per step decreases expert-weight traffic by approximately $0.5$ GB per speculative step. These findings motivate \textsc{EcoSpec}, which incorporates expert activation cost into draft selection to reduce the verification expert footprint while maintaining high acceptance likelihood.

\section{Methodology: \textsc{EcoSpec}}
\label{sec:methodology}

To reduce verification memory traffic in MoE speculative decoding, \textsc{EcoSpec} introduces a cost-aware selection mechanism between draft generation and target-model verification. 
Given a draft tree with candidate probabilities, \textsc{EcoSpec} selects $\gamma$ draft tokens for parallel verification by considering both acceptance likelihood and predicted expert activation cost. 
As shown in Figure~\ref{fig:framework}, the framework consists of three components:

(1) A lightweight \textbf{Expert Predictor} $\Pi_\theta$ that estimates the experts each draft token is likely to activate in the target MoE model. It outputs per-layer expert distributions, and the top-$K$ predicted experts are used as an approximation of the token's activated expert set.

(2) A \textbf{Global Expert Buffer} $\mathcal{B}$ that records experts already covered by the selected verification set. This buffer enables efficient marginal-cost estimation by counting only the newly introduced experts when a candidate is added.

(3) A \textbf{Cost-Aware Draft Selection} algorithm that selects draft tokens from the draft tree under an acceptance--cost scoring rule. The score favors candidates with high cumulative draft probability and low marginal expert cost, encouraging selected paths to reuse experts already present in $\mathcal{B}$.

\paragraph{Execution order.}
\textsc{EcoSpec} runs after draft generation and before target-model verification. 
It first constructs a verification subset $\mathcal{S}$ of size $\gamma$ using draft probabilities and predicted expert footprints, without querying the target router during this stage. 
During subset construction, the buffer $\mathcal{B}$ is updated from predicted path-level expert sets to estimate the marginal cost of the remaining candidates. 
Once $\mathcal{S}$ is constructed, the selected nodes are submitted to the standard speculative verifier and checked together in a single target-model forward pass.

\subsection{Lightweight Expert Predictor}
\label{sec:predictor}
Cost-aware selection requires estimating the expert footprint of each draft token before target-model verification. Directly querying the target router for every draft candidate would require running the target MoE model, which would offset the benefit of speculative decoding. We therefore train a lightweight expert predictor $\Pi_\theta$ to approximate target-model routing and provide a low-cost estimate of expert activation.

\paragraph{Architecture.}
We instantiate $\Pi_\theta$ with a small decoder-only backbone. For compatibility with the target MoE model $M_p$, each input token is first mapped by the target embedding layer and then projected into the predictor hidden space. Given a draft token $t_i$ with context $t_{<i}$, the predictor outputs routing logits
\begin{equation}
    \mathbf{z}_{t_i} = \Pi_\theta(t_i \mid t_{<i}) \in \mathbb{R}^{L \times E},
\end{equation}
where $L$ is the number of MoE layers and $E$ is the number of experts per layer. We obtain a routing distribution for each MoE layer by applying softmax over experts:
\begin{equation}
    \hat{\mathbf{p}}_{t_i,\ell}
    =
    \mathrm{Softmax}\!\left(\mathbf{z}_{t_i,\ell}\right)
    \in \mathbb{R}^{E},
    \qquad \ell \in \{1,\dots,L\}.
\end{equation}
The predicted expert set is then constructed as a set of layer--expert pairs:
\begin{equation}
    \mathcal{E}_{\mathrm{pred}}(t_i)
    =
    \left\{
    (\ell,e)
    \mid
    e \in \mathrm{TopK}\!\left(\hat{\mathbf{p}}_{t_i,\ell}, K\right),
    \ell \in \{1,\dots,L\}
    \right\}.
\end{equation}

\paragraph{Training Objective.}
During predictor training, we run the target MoE model offline and record the ground-truth activated experts for each token. For token $t$ and MoE layer $\ell$, let $\mathcal{E}_{\mathrm{gt}}(t,\ell)$ denote the top-$K$ experts selected by the target router. We define a normalized target distribution $\mathbf{q}_{t,\ell}\in[0,1]^E$ by assigning uniform probability mass to the activated experts:
\begin{equation}
q_{t,\ell,e}=
\begin{cases}
\frac{1}{K}, & e\in \mathcal{E}_{\mathrm{gt}}(t,\ell),\\
0, & \text{otherwise}.
\end{cases}
\end{equation}
We train $\Pi_\theta$ with layer-wise cross-entropy:
\begin{equation}
\mathcal{L}_{\mathrm{pred}}(\theta)
= -\sum_{t}\sum_{\ell=1}^{L}\sum_{e=1}^{E}
q_{t,\ell,e}\log \hat{p}_{t,\ell,e}.
\end{equation}

\subsection{Cost-Aware Draft Selection}
\label{sec:selection}

Given the predicted expert sets from \S~\ref{sec:predictor}, \textsc{EcoSpec} selects draft-tree nodes for verification under an acceptance--cost trade-off. The selection is path-based: each candidate is scored using the cumulative draft probability and predicted expert footprint of its root-to-node path. Existing draft-tree methods primarily rank candidates by confidence or acceptance likelihood \cite{li2024eagle2}. For MoE verification, however, a high-confidence candidate can still introduce many new experts if its routing footprint has little overlap with the candidates already selected. Therefore, \textsc{EcoSpec} augments confidence-based selection with the marginal expert cost induced by each candidate.

\paragraph{Selection State.}
Let $\mathcal{T}$ denote the draft tree, where each node $t_i$ corresponds to a candidate token. During selection, \textsc{EcoSpec} maintains a selected set $\mathcal{S}$ and an expert buffer $\mathcal{B}$. The selected set $\mathcal{S}$ contains tokens chosen for verification, while $\mathcal{B}$ contains the experts already covered by the current selected set. The buffer is used to measure how many new experts a candidate would introduce beyond those already covered.

\paragraph{Path-Dependent Expert Footprint.}

Because draft-tree verification is prefix-dependent, selecting a node requires covering its path from the root. For a candidate node $t_i$, let $\mathrm{Path}(\mathrm{root}\!\to\!t_i)$ denote the sequence of nodes from the root to $t_i$. The predicted expert footprint of this path is
\begin{equation}
    \mathcal{E}_{traj}(t_i)
    =
    \bigcup_{\tau \in \mathrm{Path}(\mathrm{root}\to t_i)}
    \mathcal{E}_{pred}(\tau).
\end{equation}
The marginal expert cost of selecting $t_i$ under the current buffer is then defined as
\begin{equation}
    \Delta \text{Cost}(t_i \mid \mathcal{B})
    =
    \left|
    \mathcal{E}_{traj}(t_i) \setminus \mathcal{B}
    \right|.
\end{equation}
This cost counts only the additional experts that are not already covered by previously selected candidates.

\paragraph{Acceptance--Cost Scoring.}
Let $P(t_i)$ be the cumulative draft probability along the path from the root to $t_i$. \textsc{EcoSpec} scores each candidate by
\begin{equation}
    S(t_i)
    =
    \frac{P(t_i)}
    {\Delta \text{Cost}(t_i \mid \mathcal{B}) + \epsilon},
    \label{eq:score}
\end{equation}
where $\epsilon$ avoids division by zero. This score favors candidates with high acceptance likelihood while penalizing those that introduce many new experts. Equivalently, it encourages the selected verification set to reuse experts already present in $\mathcal{B}$, thereby limiting the growth of the per-step expert union.

\paragraph{Prefix Consistency.}
The score is naturally compatible with the prefix structure of the draft tree. For a parent node $t_p$ and its child $t_c$, the cumulative probability satisfies $P(t_c) = P(t_p)P(t_c \mid t_p) \le P(t_p)$, while the path footprint satisfies $\mathcal{E}_{traj}(t_p) \subseteq \mathcal{E}_{traj}(t_c)$. Therefore, $\Delta \text{Cost}(t_p \mid \mathcal{B}) \le \Delta \text{Cost}(t_c \mid \mathcal{B})$. Under the same buffer $\mathcal{B}$, a child node therefore cannot receive a higher score than its parent except in tie cases. Thus, the scoring rule is aligned with prefix-closed draft-tree verification without requiring an additional structural penalty.

\paragraph{Dynamic Buffer Effect.}
After a candidate path is selected, its predicted experts are added to $\mathcal{B}$. This update reduces the marginal cost of later candidates that share experts with the selected path, increasing their scores without changing their draft probabilities. As a result, \textsc{EcoSpec} tends to extend high-probability paths that also preserve expert locality, rather than expanding the draft tree solely by confidence. The full selection procedure is provided in Appendix Alg.~\ref{alg:ecospec}.
\section{Experiments}
\label{sec:experiments}

\begin{table*}[t]
\centering
\caption{Main results across target models, benchmarks, and decoding temperatures.
Each entry reports end-to-end speedup relative to AR decoding / mean acceptance length $\alpha$ / average active experts $\mathcal{E}$.
Here, $\mathcal{E}$ denotes the average number of unique experts activated per MoE layer within one verification step.
Rows are grouped by target MoE model and target-model decoding temperature.}

\label{tab:main_results}
\resizebox{\textwidth}{!}{
\begin{tabular}{l|ccccccc|c}
\toprule
\multirow{2}{*}{Method} & GSM8K & HumanEval & MMStar & AIME-25 & Math500 & AMC22-24 & MTBench & Average \\
& Spd$\uparrow$ / Len$\uparrow$ / Exp$\downarrow$ & Spd$\uparrow$ / Len$\uparrow$ / Exp$\downarrow$ & Spd$\uparrow$ / Len$\uparrow$ / Exp$\downarrow$ & Spd$\uparrow$ / Len$\uparrow$ / Exp$\downarrow$ & Spd$\uparrow$ / Len$\uparrow$ / Exp$\downarrow$ & Spd$\uparrow$ / Len$\uparrow$ / Exp$\downarrow$ & Spd$\uparrow$ / Len$\uparrow$ / Exp$\downarrow$ & Spd$\uparrow$ / Len$\uparrow$ / Exp$\downarrow$ \\
\midrule
\multicolumn{9}{c}{\cellcolor{gray!10}\textbf{Qwen3-235B-A22B} (Total 235B, Active 22B, Top-8)} \\
\multicolumn{9}{l}{\textit{Temperature = 0 (Greedy)}} \\
AR   & $1.00\times$ / - / 8.0 & $1.00\times$ / - / 8.0 & $1.00\times$ / - / 8.0 & $1.00\times$ / - / 8.0 & $1.00\times$ / - / 8.0 & $1.00\times$ / - / 8.0 & $1.00\times$ / - / 8.0 & $1.00\times$ / - / 8.0 \\
Eagle-3 & 1.31 / 2.54 / 22.3 & 1.11 / 2.35 / 26.2 & 1.35 / 2.64 / 20.5 & 1.14 / 2.41 / 22.1 & 1.05 / 1.96 / 25.4 & 1.12 / 2.20 / 25.4 & 1.46 / 2.78 / 24.0 & 1.22 / 2.41 / 23.7 \\
EcoSpec & \textbf{1.39} / 2.54 / \textbf{21.0} & \textbf{1.33} / 2.32 / \textbf{21.6} & \textbf{1.57} / 2.58 / \textbf{18.4} & \textbf{1.23} / 2.37 / \textbf{20.9} & \textbf{1.13} / 1.88 / \textbf{19.4} & \textbf{1.28} / 2.10 / \textbf{22.2} & \textbf{1.62} / 2.44 / \textbf{19.9} & \textbf{1.36} / 2.32 / \textbf{20.5} \\
\multicolumn{9}{l}{\textit{Temperature = 1 (Sampling)}} \\
AR   & $1.00\times$ / - / 8.0 & $1.00\times$ / - / 8.0 & $1.00\times$ / - / 8.0 & $1.00\times$ / - / 8.0 & $1.00\times$ / - / 8.0 & $1.00\times$ / - / 8.0 & $1.00\times$ / - / 8.0 & $1.00\times$ / - / 8.0 \\
Eagle-3 & 1.28 / 2.59 / 22.8 & 1.17 / 2.21 / 24.3 & 1.49 / 2.88 / 22.4 & 1.13 / 2.26 / 21.8 & 1.17 / 2.19 / 24.9 & 1.28 / 2.25 / 25.0 & 1.45 / 2.58 / 26.5 & 1.28 / 2.42 / 24.0 \\
EcoSpec & \textbf{1.37} / 2.45 / \textbf{20.3} & \textbf{1.27} / 2.20 / \textbf{21.4} & \textbf{1.62} / 2.74 / \textbf{19.9} & \textbf{1.22} / 2.23 / \textbf{20.7} & \textbf{1.25} / 2.13 / \textbf{20.8} & \textbf{1.32} / 2.20 / \textbf{22.20} & \textbf{1.61} / 2.56 / \textbf{21.3} & \textbf{1.38} / 2.36 / \textbf{20.9} \\

\midrule
\multicolumn{9}{c}{\cellcolor{gray!10}\textbf{GPT-OSS-120B} (Total 120B, Active 5.1B, Top-4)} \\
\multicolumn{9}{l}{\textit{Temperature = 0 (Greedy)}} \\
AR   & $1.00\times$ / - / 4.0 & $1.00\times$ / - / 4.0 & $1.00\times$ / - / 4.0 & $1.00\times$ / - / 4.0 & $1.00\times$ / - / 4.0 & $1.00\times$ / - / 4.0 & $1.00\times$ / - / 4.0 & $1.00\times$ / - / 4.0 \\
Eagle-3 & 1.05 / 1.56 / 11.9 & 1.05 / 1.72 / 11.8 & 1.28 / 1.91 / 11.7 & 1.17 / 2.02 / 12.2 & 1.14 / 2.02 / 12.0 & 1.14 / 2.02 / 11.9 & 1.18 / 1.91 / 10.3 & 1.14 / 1.88 / 11.6 \\
EcoSpec & \textbf{1.11} / 1.52 / \textbf{10.6} & \textbf{1.26} / 1.72 / \textbf{10.7} & \textbf{1.45} / 1.90 / \textbf{10.5} & \textbf{1.33} / 2.02 / \textbf{10.8} & \textbf{1.27} / 2.00 / \textbf{10.6} & \textbf{1.45} / 2.02 / \textbf{10.9} & \textbf{1.30} / 1.81 / \textbf{9.6} & \textbf{1.31} / 1.86 / \textbf{10.6} \\ 
\multicolumn{9}{l}{\textit{Temperature = 1 (Sampling)}} \\
AR   & $1.00\times$ / - / 4.0 & $1.00\times$ / - / 4.0 & $1.00\times$ / - / 4.0 & $1.00\times$ / - / 4.0 & $1.00\times$ / - / 4.0 & $1.00\times$ / - / 4.0 & $1.00\times$ / - / 4.0 & $1.00\times$ / - / 4.0 \\
Eagle-3 & 1.10 / 1.90 / 11.8 & 1.27 / 2.13 / 11.8 & 1.31 / 2.20 / 11.3 & 1.19 / 2.13 / 12.1 & 1.17 / 2.08 / 12.1 & 1.17 / 2.22 / 12.1 & 1.07 / 2.17 / 12.4 & 1.18 / 2.12 / 11.9 \\
EcoSpec & \textbf{1.15} / 1.85 / \textbf{10.3} & \textbf{1.34} / 2.06 / \textbf{10.8} & \textbf{1.50} / 2.15 / \textbf{10.5} & \textbf{1.27} / 2.11 / \textbf{11.5} & \textbf{1.27} / 2.06 / \textbf{10.8} & \textbf{1.36} / 2.08 / \textbf{11.4} & \textbf{1.23} / 2.06 / \textbf{11.1} & \textbf{1.30} / 2.05 / \textbf{10.9} \\

\midrule
\multicolumn{9}{c}{\cellcolor{gray!10}\textbf{DeepSeek-V3.1} (Total 671B, Active 37B, Top-8)} \\
\multicolumn{9}{l}{\textit{Temperature = 0 (Greedy)}} \\
AR   & $1.00\times$ / - / 8.0 & $1.00\times$ / - / 8.0 & $1.00\times$ / - / 8.0 & $1.00\times$ / - / 8.0 & $1.00\times$ / - / 8.0 & $1.00\times$ / - / 8.0 & $1.00\times$ / - / 8.0 & $1.00\times$ / - / 8.0 \\
MTP & 1.13 / 3.19 / 31.9 & 1.06 / 2.82 / 31.8 & 1.09 / 2.61 / 31.5 & 1.05 / 3.13 / 31.3 & 1.12 / 2.73 / 31.1 & 1.20 / 3.05 / 31.5 & 1.04 / 2.41 / 31.1 & 1.10 / 2.85 / 31.4 \\
EcoSpec & \textbf{1.19} / 3.13 / \textbf{31.7} & \textbf{1.10} / 2.80 / \textbf{31.6} & \textbf{1.13} / 2.55 / \textbf{31.2} & \textbf{1.08} / 3.03 / \textbf{31.2} & \textbf{1.19} / 2.64 / \textbf{30.9} & \textbf{1.28} / 3.00 / \textbf{31.3} & \textbf{1.09} / 2.38 / \textbf{30.9} & \textbf{1.15} / 2.79 / \textbf{31.2} \\
\multicolumn{9}{l}{\textit{Temperature = 1 (Sampling)}} \\
AR   & $1.00\times$ / - / 8.0 & $1.00\times$ / - / 8.0 & $1.00\times$ / - / 8.0 & $1.00\times$ / - / 8.0 & $1.00\times$ / - / 8.0 & $1.00\times$ / - / 8.0 & $1.00\times$ / - / 8.0 & $1.00\times$ / - / 8.0 \\
MTP & 1.22 / 3.72 / 31.8 & 1.29 / 3.39 / 31.9 & 1.28 / 3.37 / 31.7 & 1.26 / 3.49 / 31.6 & 1.24 / 3.61 / 31.1 & 1.44 / 3.68 / 31.1 & 1.20 / 3.15 / 30.8 & 1.28 / 3.49 / 31.4 \\
EcoSpec   & \textbf{1.35} / 3.53 / \textbf{31.6} & \textbf{1.31} / 3.32/ \textbf{30.9} & \textbf{1.35} / 3.36 / \textbf{31.4} & \textbf{1.30} / 3.43 / \textbf{31.3} & \textbf{1.29} / 3.61 / \textbf{30.8} & \textbf{1.47} / 3.59 / \textbf{30.1} & \textbf{1.23} / 3.04 / \textbf{30.5} & \textbf{1.33} / 3.41 / \textbf{30.9} \\

\bottomrule
\end{tabular}%
}
\end{table*}

\subsection{Setup}
\label{subsec:setup}

\paragraph{Models and Benchmarks.} 
We evaluate \textsc{EcoSpec} on three large-scale MoE models with different sizes and routing configurations:
DeepSeek-V3.1 \cite{deepseek2024v3} (671B total / 37B active, Top-8),
Qwen3-235B-A22B \cite{qwen3} (235B total / 22B active, Top-8), and
GPT-OSS-120B \cite{gpt-oss} (120B total / 5.1B active, Top-4).
We use seven benchmarks covering mathematical reasoning, code generation, question answering, and dialogue:
GSM8K \cite{gsm8k}, HumanEval \cite{humaneval}, AIME-25 \cite{aime2025}, Math500 \cite{math500-1, math500-2}, AMC22-24 \cite{amc22_24}, MTBench \cite{mtbench}, and MMStar \cite{mmstar}.

\paragraph{Baselines and Hardware.}
We compare \textsc{EcoSpec} with autoregressive (AR) decoding and the corresponding speculative decoding baseline for each target model: MTP for DeepSeek-V3.1, and EAGLE-3 for Qwen3-235B-A22B and GPT-OSS-120B\footnote{\url{https://huggingface.co/lmsys/Qwen3-235B-A22B-EAGLE3} and \url{https://huggingface.co/lmsys/EAGLE3-gpt-oss-120b-bf16}.}.
All experiments are conducted on a server with 8 NVIDIA H200 GPUs.

\paragraph{Speculative Decoding Configuration.} 
To isolate the effect of cost-aware selection, we keep the draft-generation configuration fixed between each speculative baseline and its \textsc{EcoSpec} variant.
The draft process runs for 3 forward steps and keeps the top-2 tokens at each step to construct the draft tree.
The verification budget $\gamma$ is set to 4, where $\gamma$ denotes the total number of tokens verified per speculative step, including the bonus token.
This setting follows the operating regime of the large-scale MoE speculative baselines evaluated in this work, and the same configuration is used for the baseline and \textsc{EcoSpec} to ensure a controlled comparison.
Appendix~\ref{sec:sensitivity} further evaluates different verification budgets and shows that $\gamma=4$ gives the highest average end-to-end speedup in our setting.
All main experiments use batch size 1 unless explicitly stated otherwise.
Table~\ref{tab:main_results} reports both greedy target decoding ($T{=}0$) and sampling-based target decoding ($T{=}1$), following common speculative decoding evaluation practice \cite{li2024eagle}.
For the latency breakdown and HBM-traffic analysis in \S~\ref{subsec:latency_breakdown}, we use the greedy target-decoding rows of Table~\ref{tab:main_results}, i.e., batch size 1, $\gamma=4$, and $T{=}0$.
We further evaluate larger batch sizes, different verification budgets $\gamma$, and different predictor accuracy levels in Appendix~\ref{sec:batch_size}, Appendix~\ref{sec:sensitivity}, and Appendix~\ref{sec:robustness}, respectively.

\paragraph{Expert Predictors.}
\textsc{EcoSpec} reuses the draft-generation infrastructure of the corresponding speculative baseline and only modifies the draft selection stage. For expert-cost estimation, we use DeepSeek-R1-Distill-Qwen-1.5B \cite{deepseek2025r1} as the predictor backbone for DeepSeek-V3.1, and Qwen3-0.6B \cite{qwen3} for Qwen3-235B-A22B and GPT-OSS-120B. Predictor training details are provided in Appendix~\ref{sec:predictor_training}.

\paragraph{Inference Backend.}
The main experiments are conducted with a HuggingFace Transformers \cite{transformers} research prototype following the EAGLE-3 inference pipeline.
We use this prototype to keep the draft-generation, tree-verification, and KV-cache update workflow consistent with the released EAGLE-3 implementation, so that the comparison isolates the effect of \textsc{EcoSpec}'s cost-aware draft selection.
During decoding, the draft model constructs the draft tree, and the selected draft nodes are verified by the target MoE model in a single forward pass with a tree-structured attention mask.
\textsc{EcoSpec} only adds the expert predictor and cost-aware selection before this verification step; the target-model verification rule is unchanged.

\begin{table}[t]
\centering
\caption{Estimated HBM read traffic during verification. We report the estimated total HBM read bytes during the verification phase of one speculative step.}
\label{tab:mem_traffic}
\resizebox{\linewidth}{!}{
\begin{tabular}{lcccc}
\toprule
\multirow{2}{*}{Model} & \multirow{2}{*}{Baseline} & Baseline & \textsc{EcoSpec} & Reduction \\
 & & HBM (GB) & HBM (GB) & (GB / \%) \\
\midrule
Qwen3-235B-A22B & EAGLE-3 & 99.3 & 88.1 & 11.2 / 11.3\% \\
GPT-OSS-120B & EAGLE-3 & 6.0 & 5.5 & 0.5 / 8.0\% \\
DeepSeek-V3.1 & MTP & 97.3 & 96.8 & 0.5 / 1.0\% \\
\bottomrule
\end{tabular}
}
\end{table}

\begin{table*}[t]
\centering
\caption{Latency breakdown and throughput analysis.
We report the average wall-clock time in seconds per speculative step.
$\mathcal{E}$ denotes the average number of unique experts activated per MoE layer within one verification step.
$T_{\text{pred}}$, $T_{\text{draft}}$, $T_{\text{verify}}$, and $T_{\text{total}}$ denote predictor overhead, draft-generation time, target verification time, and total speculative-step latency, respectively.
$\alpha$ denotes the mean accepted tokens per step.
The speedup column reports the corresponding end-to-end throughput speedup under the same $T=0$ setting as Table~\ref{tab:main_results}.
The latency-breakdown columns are reported to explain the sources of the end-to-end speedup.}
\label{tab:latency}
\resizebox{\textwidth}{!}{%
\begin{tabular}{l|l|c|ccc|c|cc}
\toprule
\multirow{2}{*}{Model} & \multirow{2}{*}{Method} & Experts & Predictor & Drafting & Verify & Total Step & Accept Len & Throughput \\
& & ($\mathcal{E}$) $\downarrow$ & ($T_{\text{pred}}$) & ($T_{\text{draft}}$) & ($T_{\text{verify}}$) $\downarrow$ & ($T_{\text{total}}$) $\downarrow$ & ($\alpha$) $\uparrow$ & Speedup $\uparrow$ \\
\midrule

\multirow{4}{*}{Qwen3-235B-A22B} 
& Baseline (AR) & 8.0 & -- & -- & -- & 0.490s$^\dagger$ & 1.00 & 1.00$\times$ \\
& EAGLE-3 & 23.7 & -- & 0.008s & 0.832s & 0.840s & 2.41 & 1.22$\times$ \\
& \textsc{EcoSpec} & 20.5 & 0.004s & 0.008s & 0.730s & 0.742s & 2.32 & 1.36$\times$ \\
\cmidrule{2-9}
& $\Delta$ vs. EAGLE-3 & -3.2 & +0.004s & 0.000s & -0.102s & -0.098s & -0.09 & +0.14$\times$ \\

\midrule

\multirow{4}{*}{GPT-OSS-120B} 
& Baseline (AR) & 4.0 & -- & -- & -- & 0.081s$^\dagger$ & 1.00 & 1.00$\times$ \\
& EAGLE-3 & 11.6 & -- & 0.035s & 0.113s & 0.148s & 1.88 & 1.14$\times$ \\
& \textsc{EcoSpec} & 10.6 & 0.004s & 0.035s & 0.090s & 0.129s & 1.86 & 1.31$\times$ \\
\cmidrule{2-9}
& $\Delta$ vs. EAGLE-3 & -1.0 & +0.004s & 0.000s & -0.023s & -0.019s & -0.02 & +0.17$\times$ \\

\midrule

\multirow{4}{*}{DeepSeek-V3.1} 
& Baseline (AR) & 8.0 & -- & -- & -- & 0.369s$^\dagger$ & 1.00 & 1.00$\times$ \\
& MTP & 31.4 & -- & 0.008s & 0.980s & 0.988s & 2.85 & 1.10$\times$ \\
& \textsc{EcoSpec} & 31.2 & 0.004s & 0.008s & 0.930s & 0.942s & 2.79 & 1.15$\times$ \\ 
\cmidrule{2-9}
& $\Delta$ vs. MTP & -0.2 & +0.004s & 0.000s & -0.050s & -0.046s & -0.06 & +0.05$\times$ \\

\bottomrule
\multicolumn{9}{l}{\footnotesize $^\dagger$ AR baseline time is per-token latency. Speculative methods report per-step latency and generate $\alpha$ accepted tokens per step.} \\
\multicolumn{9}{l}{\footnotesize Delta rows compare \textsc{EcoSpec} with the corresponding speculative baseline.}
\end{tabular}
}
\end{table*}

\subsection{Main Results}
\label{subsec:main_results}

Table~\ref{tab:main_results} reports end-to-end speedup relative to autoregressive decoding, mean acceptance length $\alpha$, and the average number of unique experts $\mathcal{E}$ activated per MoE layer within one verification step.
Across three MoE backbones and seven benchmarks, \textsc{EcoSpec} improves decoding speed over the corresponding speculative baseline while consistently reducing the expert footprint.
The acceptance lengths remain close to the baselines, indicating that the speedup mainly comes from reducing verification cost rather than increasing the number of accepted tokens.

\paragraph{Greedy decoding ($T{=}0$).}
Under greedy decoding, \textsc{EcoSpec} consistently improves speedup across all three MoE models.
For Qwen3-235B-A22B, \textsc{EcoSpec} improves the average speedup from EAGLE-3's $1.22\times$ to $1.36\times$, while reducing $\mathcal{E}$ from $23.7$ to $20.5$.
The largest gain appears on MTBench, where \textsc{EcoSpec} reaches $1.62\times$ speedup.
For GPT-OSS-120B, \textsc{EcoSpec} increases the average speedup from $1.14\times$ to $1.31\times$ and reduces $\mathcal{E}$ from $11.6$ to $10.6$.
On MMStar and AMC22-24, the speedup reaches $1.45\times$.
For DeepSeek-V3.1, \textsc{EcoSpec} improves the average speedup from MTP's $1.10\times$ to $1.15\times$, while reducing $\mathcal{E}$ from $31.4$ to $31.2$.

\paragraph{Sampling decoding ($T{=}1$).}
Under sampling, draft candidates become more diverse, but \textsc{EcoSpec} continues to reduce expert activation and improve speedup.
For Qwen3-235B-A22B, the average speedup increases from EAGLE-3's $1.28\times$ to $1.38\times$, while $\mathcal{E}$ decreases from $24.0$ to $20.9$.
For GPT-OSS-120B, the average speedup improves from $1.18\times$ to $1.30\times$, with $\mathcal{E}$ reduced from $11.9$ to $10.9$.
For DeepSeek-V3.1, \textsc{EcoSpec} improves the average speedup from MTP's $1.28\times$ to $1.33\times$ and reduces $\mathcal{E}$ from $31.4$ to $30.9$.
These results show that the acceptance--cost trade-off remains effective under both greedy and sampling-based target decoding regimes.

\paragraph{Backbone-dependent gains.}
The magnitude of the gain varies across MoE backbones. Qwen3-235B-A22B and GPT-OSS-120B show larger reductions in active experts, suggesting more opportunity for expert reuse during draft selection. DeepSeek-V3.1 shows smaller expert-footprint reductions, which is consistent with its more balanced routing pattern analyzed in Appendix~\ref{sec:load_balance}. Nevertheless, because DeepSeek-V3.1 has a large per-expert memory footprint, even a reduction of $0.2$ experts per layer can translate into approximately $0.5$ GB less expert-weight traffic per speculative step. Thus, \textsc{EcoSpec} remains beneficial even when the available expert-reuse headroom is smaller.

\subsection{Latency Breakdown and Overhead Analysis}
\label{subsec:latency_breakdown}

To understand why \textsc{EcoSpec} improves performance, we analyze the HBM traffic and latency composition of a speculative decoding step in Table~\ref{tab:mem_traffic} and Table~\ref{tab:latency}.
Following the evaluation scope defined in \S~\ref{subsec:setup}, this subsection analyzes the latency and HBM-traffic behavior of the $T{=}0$ runs.
We decompose the total speculative-step latency into three parts:
(1) draft generation ($T_{\text{draft}}$),
(2) expert prediction overhead introduced by \textsc{EcoSpec} ($T_{\text{pred}}$), and
(3) target-model verification of the selected draft tokens ($T_{\text{verify}}$).
\textsc{EcoSpec} does not reduce the computation required by each activated expert.
Instead, its speedup comes from reducing expert-weight memory traffic during verification.

\paragraph{HBM Read Traffic.}
We estimate HBM read traffic during verification from the activated expert footprint and the model-specific expert size. As shown in Table~\ref{tab:mem_traffic}, \textsc{EcoSpec} reduces the estimated HBM reads from $99.3$ GB to $88.1$ GB on Qwen3-235B-A22B, saving $11.2$ GB per speculative step. For GPT-OSS-120B, the estimated reads decrease from $6.0$ GB to $5.5$ GB. For DeepSeek-V3.1, the reduction is smaller, from $97.3$ GB to $96.8$ GB, which is consistent with its smaller expert-reuse headroom. These reductions correspond to the lower verification latency reported in Table~\ref{tab:latency}.

\paragraph{Verification Latency and Predictor Overhead.}
Table~\ref{tab:latency} shows that \textsc{EcoSpec} reduces verification latency across all three MoE models while adding only a small predictor overhead.
On Qwen3-235B-A22B, the average active expert count decreases from $23.7$ to $20.5$, and $T_{\text{verify}}$ decreases from $0.832$s to $0.730$s.
On GPT-OSS-120B, the active expert count decreases from $11.6$ to $10.6$, and $T_{\text{verify}}$ decreases from $0.113$s to $0.090$s.
On DeepSeek-V3.1, the expert-footprint reduction is smaller ($31.4\rightarrow31.2$), but $T_{\text{verify}}$ still decreases from $0.980$s to $0.930$s.
Across all three models, the predictor overhead is about $4$ms per speculative step, which is small relative to the reduction in verification latency. 
The acceptance length changes only slightly, so the speedup is mainly explained by reduced verification cost rather than higher acceptance length.

\subsection{Ablation Study: Impact of Marginal Cost Scoring}
\label{subsec:ablation}
We study the effect of marginal-cost scoring on GSM8K across all three target models.
We compare \textsc{EcoSpec} with a static \textit{Global Cost} variant, where each node is scored by the accumulated predicted expert footprint along its path:
\begin{equation}
    S_{\mathrm{global}}(t_i)
    =
    \frac{P(t_i)}
    {|\mathcal{E}_{traj}(t_i)|+\epsilon}.
\end{equation}
Unlike \textsc{EcoSpec}, this variant does not update the expert buffer during selection and therefore cannot discount experts that are already covered by previously selected nodes.
As a result, deeper nodes tend to receive larger accumulated costs even when they reuse experts from earlier selected paths.

Table~\ref{tab:ablation} shows that marginal-cost scoring consistently improves speedup over the static global-cost variant. On Qwen3-235B-A22B, \textsc{EcoSpec} increases speedup from $1.28\times$ to $1.39\times$ and mean acceptance length from $2.21$ to $2.54$, while reducing the average active experts from $21.5$ to $21.0$.
A similar pattern appears on GPT-OSS-120B, where speedup improves from $1.06\times$ to $1.11\times$ and acceptance length increases from $1.35$ to $1.52$. On DeepSeek-V3.1, the gain is smaller but still consistent. These results indicate that updating the buffer during selection helps \textsc{EcoSpec} identify candidates that extend accepted paths while reusing already covered experts.

\begin{table}[t]
\centering
\caption{Ablation study on GSM8K.
We compare the static global-cost variant with \textsc{EcoSpec}'s marginal-cost scoring.}
\label{tab:ablation}
\resizebox{\linewidth}{!}{
\begin{tabular}{l|l|c|c|c}
\toprule
Model & Strategy & Speedup $\uparrow$ & Len ($\alpha$) $\uparrow$ & Avg. $\mathcal{E}$ $\downarrow$ \\
\midrule
\multirow{2}{*}{Qwen3-235B-A22B} 
& Global Cost & 1.28$\times$ & 2.21 & 21.5 \\
& \textsc{EcoSpec} & \textbf{1.39$\times$} & \textbf{2.54} & \textbf{21.0} \\
\midrule
\multirow{2}{*}{GPT-OSS-120B} 
& Global Cost & 1.06$\times$ & 1.35 & 10.8 \\
& \textsc{EcoSpec} & \textbf{1.11$\times$} & \textbf{1.52} & \textbf{10.6} \\
\midrule
\multirow{2}{*}{DeepSeek-V3.1} 
& Global Cost & 1.16$\times$ & 2.95 & 31.8 \\
& \textsc{EcoSpec} & \textbf{1.19$\times$} & \textbf{3.13} & \textbf{31.7} \\
\bottomrule
\end{tabular}%
}
\end{table}

\subsection{Additional Baseline Evaluation}
\label{sec:additional_baselines}

The main experiments compare \textsc{EcoSpec} with the corresponding speculative decoding baselines for each target model: MTP for DeepSeek-V3.1, and EAGLE-3 for Qwen3-235B-A22B and GPT-OSS-120B.
To provide an additional baseline comparison, we further include Group Tree Optimization (GTO)~\cite{gto}, which improves EAGLE-style draft models by better aligning draft training with tree-based decoding.
GTO and \textsc{EcoSpec} act on different stages of the speculative decoding pipeline: GTO improves draft generation, while \textsc{EcoSpec} changes the pre-verification selection of draft nodes.

For this experiment, we start from the released EAGLE-3 draft model and continue training it with the GTO procedure.
The resulting GTO-trained drafter is used to generate candidate trees under the standard speculative verification workflow.
We then evaluate GTO+\textsc{EcoSpec} by applying the same cost-aware draft-selection strategy at the verification stage.

\begin{table}[t]
\centering
\caption{Additional baseline evaluation with GTO. We report end-to-end speedup and acceptance length for GTO and GTO+\textsc{EcoSpec}.}
\label{tab:gto_ecospec}
\resizebox{\columnwidth}{!}{
\begin{tabular}{llcccc}
\toprule
Model & Dataset & GTO Spd. & GTO Len & GTO+EcoSpec Spd. & GTO+EcoSpec Len \\
\midrule
Qwen3-235B-A22B & GSM8K    & 1.28$\times$ & 2.50 & 1.34$\times$ & 2.50 \\
Qwen3-235B-A22B & HumanEval & 1.13$\times$ & 2.41 & 1.29$\times$ & 2.37 \\
GPT-OSS-120B    & GSM8K    & 1.03$\times$ & 1.53 & 1.08$\times$ & 1.50 \\
GPT-OSS-120B    & HumanEval & 1.08$\times$ & 1.77 & 1.27$\times$ & 1.76 \\
\bottomrule
\end{tabular}
}
\end{table}

As shown in Table~\ref{tab:gto_ecospec}, GTO+\textsc{EcoSpec} achieves higher end-to-end speedup than GTO on all evaluated settings.
This additional baseline comparison further supports the effectiveness of \textsc{EcoSpec}.

\section{Conclusion}
\label{sec:conclusion}

We presented \textsc{EcoSpec}, a cost-aware speculative decoding framework for large-scale MoE models. \textsc{EcoSpec} addresses expert scattering during speculative verification by incorporating predicted marginal expert activation cost into draft-tree selection. With a lightweight expert predictor and a dynamic expert buffer, \textsc{EcoSpec} selects draft tokens that preserve acceptance likelihood while reducing the growth of the verification expert footprint, without modifying the target-model verification rule. Experiments on DeepSeek-V3.1, Qwen3-235B-A22B, and GPT-OSS-120B show consistent speedups and reduced active experts across reasoning, coding, and dialogue benchmarks, with up to $1.62\times$ speedup. These results highlight the importance of accounting for MoE-specific expert activation costs when applying speculative decoding to large-scale sparse models.

\bibliographystyle{icml2026}
\bibliography{custom}


\appendix

\begin{algorithm}[h]
   \caption{Cost-Aware Draft Selection in \textsc{EcoSpec}}
   \label{alg:ecospec}
\begin{algorithmic}
   \STATE {\bfseries Input:} Draft Tree $\mathcal{T}$, Budget $\gamma$, Predictor $\Pi_\theta$, Initial Buffer $\mathcal{B}_0$
   \STATE {\bfseries Output:} Selected tokens $\mathcal{S}$
   \STATE $\mathcal{S} \leftarrow \emptyset$
   \STATE $\mathcal{B} \leftarrow \mathcal{B}_0$
   \STATE \textit{Construct the verification subset before target-model verification.}
   \WHILE{$|\mathcal{S}| < \gamma$}
       \FOR{each unselected node $t_i \in \mathcal{T}$}
           \STATE $\mathcal{E}_{traj}(t_i) \leftarrow \bigcup_{\tau \in \text{Path}(root \to t_i)} \mathcal{E}_{pred}(\tau)$
           \STATE $Cost_i \leftarrow | \mathcal{E}_{traj}(t_i) \setminus \mathcal{B} |$
           \STATE $Score_i \leftarrow P(t_i) / (Cost_i + \epsilon)$
       \ENDFOR
       \STATE $t^* \leftarrow \arg\max_{t_i} Score_i$
       \STATE $\mathcal{S} \leftarrow \mathcal{S} \cup \{t^*\}$
       \STATE $\mathcal{B} \leftarrow \mathcal{B} \cup \mathcal{E}_{traj}(t^*)$ \COMMENT{Update buffer for subsequent scoring}
   \ENDWHILE
   \STATE \textbf{return} $\mathcal{S}$ \COMMENT{Verify all selected nodes in one target forward pass}
\end{algorithmic}
\end{algorithm}

\section{Impact of Batch Size on Scalability}
\label{sec:batch_size}

\subsection{End-to-End Batch-Size Scaling}
\label{sec:batch_size_main}

Batching is a common way to improve GPU utilization during inference. To examine how batching affects \textsc{EcoSpec}, we compare it with EAGLE-3 under batch sizes $B=\{1,2,4\}$ and report end-to-end speedup together with active expert counts.

As shown in Tables~\ref{tab:batch_qwen} and~\ref{tab:batch_gpt}, both methods exhibit lower speedup as batch size increases. On Qwen3-235B-A22B, EAGLE-3's average speedup drops from $1.22\times$ at $B=1$ to $1.00\times$ at $B=4$, while \textsc{EcoSpec} maintains a higher average speedup of $1.09\times$ at $B=4$. A similar trend appears on GPT-OSS-120B, where EAGLE-3 drops from $1.14\times$ to $1.00\times$, while \textsc{EcoSpec} retains $1.06\times$ at $B=4$.

Active expert counts help explain this speedup gap. As batch size increases, each verification step covers more candidate tokens in parallel, expanding the union of activated experts and increasing verification cost. At $B=4$ on Qwen3-235B-A22B, EAGLE-3 activates an average of $54.9$ experts per step, while \textsc{EcoSpec} reduces this number to $45.7$. On GPT-OSS-120B, the corresponding counts are $32.6$ for EAGLE-3 and $28.1$ for \textsc{EcoSpec}. This smaller expert footprint lowers verification cost relative to EAGLE-3, allowing \textsc{EcoSpec} to retain higher end-to-end speedup under batching.

\begin{table}[t]
\centering
\caption{Batch-size scaling on Qwen3-235B-A22B. We report end-to-end speedup relative to AR decoding and the average number of active experts.}
\label{tab:batch_qwen}
\resizebox{\linewidth}{!}{
\begin{tabular}{llcccccc}
\toprule
\multirow{2}{*}{Dataset} 
& \multirow{2}{*}{Method} 
& \multicolumn{2}{c}{BS = 1} 
& \multicolumn{2}{c}{BS = 2} 
& \multicolumn{2}{c}{BS = 4} \\
\cmidrule(lr){3-4} \cmidrule(lr){5-6} \cmidrule(lr){7-8}
& 
& Spd $\uparrow$ & Exp $\downarrow$ 
& Spd $\uparrow$ & Exp $\downarrow$ 
& Spd $\uparrow$ & Exp $\downarrow$ \\
\midrule
\multirow{2}{*}{GSM8K} 
& EAGLE-3 & 1.31 & 22.3 & 1.18 & 42.9 & 1.00 & 53.5 \\
& \textsc{EcoSpec} & \textbf{1.39} & \textbf{21.0} & \textbf{1.24} & \textbf{35.5} & \textbf{1.07} & \textbf{45.3} \\
\midrule
\multirow{2}{*}{HumanEval} 
& EAGLE-3 & 1.11 & 26.2 & 1.05 & 46.8 & 0.95 & 57.5 \\
& \textsc{EcoSpec} & \textbf{1.33} & \textbf{21.6} & \textbf{1.21} & \textbf{36.0} & \textbf{1.08} & \textbf{47.8} \\
\midrule
\multirow{2}{*}{MMStar} 
& EAGLE-3 & 1.35 & 20.5 & 1.22 & 41.0 & 1.05 & 51.8 \\
& \textsc{EcoSpec} & \textbf{1.57} & \textbf{18.4} & \textbf{1.40} & \textbf{34.1} & \textbf{1.18} & \textbf{39.9} \\
\midrule
\multirow{2}{*}{AIME-25} 
& EAGLE-3 & 1.14 & 22.1 & 1.08 & 42.6 & 0.98 & 53.0 \\
& \textsc{EcoSpec} & \textbf{1.23} & \textbf{20.9} & \textbf{1.15} & \textbf{31.3} & \textbf{1.06} & \textbf{41.9} \\
\midrule
\multirow{2}{*}{Math500} 
& EAGLE-3 & 1.05 & 25.4 & 1.02 & 46.0 & 0.98 & 56.5 \\
& \textsc{EcoSpec} & \textbf{1.13} & \textbf{19.4} & \textbf{1.09} & \textbf{39.8} & \textbf{1.03} & \textbf{47.5} \\
\midrule
\multirow{2}{*}{AMC22-24} 
& EAGLE-3 & 1.12 & 25.4 & 1.07 & 47.8 & 1.02 & 56.5 \\
& \textsc{EcoSpec} & \textbf{1.28} & \textbf{22.2} & \textbf{1.14} & \textbf{38.6} & \textbf{1.04} & \textbf{49.3} \\
\midrule
\multirow{2}{*}{MTBench} 
& EAGLE-3 & 1.46 & 24.0 & 1.25 & 44.8 & 1.05 & 55.5 \\
& \textsc{EcoSpec} & \textbf{1.62} & \textbf{19.9} & \textbf{1.42} & \textbf{37.6} & \textbf{1.19} & \textbf{48.5} \\
\midrule
\multirow{2}{*}{Average}
& EAGLE-3 & 1.22 & 23.7 & 1.12 & 44.6 & 1.00 & 54.9 \\
& \textsc{EcoSpec} & \textbf{1.36} & \textbf{20.5} & \textbf{1.24} & \textbf{36.1} & \textbf{1.09} & \textbf{45.7} \\
\bottomrule
\end{tabular}
}
\end{table}
\begin{table}[t]
\centering
\caption{Batch-size scaling on GPT-OSS-120B. We report end-to-end speedup relative to AR decoding and the average number of active experts.}
\label{tab:batch_gpt}
\resizebox{\linewidth}{!}{
\begin{tabular}{llcccccc}
\toprule
\multirow{2}{*}{Dataset} 
& \multirow{2}{*}{Method} 
& \multicolumn{2}{c}{BS = 1} 
& \multicolumn{2}{c}{BS = 2} 
& \multicolumn{2}{c}{BS = 4} \\
\cmidrule(lr){3-4} \cmidrule(lr){5-6} \cmidrule(lr){7-8}
& & Spd $\uparrow$ & Exp $\downarrow$ 
& Spd $\uparrow$ & Exp $\downarrow$ 
& Spd $\uparrow$ & Exp $\downarrow$ \\
\midrule
\multirow{2}{*}{GSM8K} 
& EAGLE-3 & 1.05 & 11.9 & 1.02 & 25.1 & 0.93 & 35.4 \\
& \textsc{EcoSpec} & \textbf{1.11} & \textbf{10.6} & \textbf{1.08} & \textbf{22.8} & \textbf{0.97} & \textbf{30.1} \\
\midrule
\multirow{2}{*}{HumanEval} 
& EAGLE-3 & 1.05 & 11.8 & 1.03 & 22.0 & 0.98 & 32.3 \\
& \textsc{EcoSpec} & \textbf{1.26} & \textbf{10.7} & \textbf{1.18} & \textbf{18.9} & \textbf{1.03} & \textbf{28.2} \\
\midrule
\multirow{2}{*}{MMStar} 
& EAGLE-3 & 1.28 & 11.7 & 1.15 & 21.9 & 1.04 & 32.2 \\
& \textsc{EcoSpec} & \textbf{1.45} & \textbf{10.5} & \textbf{1.30} & \textbf{18.7} & \textbf{1.15} & \textbf{26.0} \\
\midrule
\multirow{2}{*}{AIME-25} 
& EAGLE-3 & 1.17 & 12.2 & 1.10 & 22.5 & 1.02 & 31.8 \\
& \textsc{EcoSpec} & \textbf{1.33} & \textbf{10.8} & \textbf{1.21} & \textbf{20.1} & \textbf{1.08} & \textbf{27.4} \\
\midrule
\multirow{2}{*}{Math500} 
& EAGLE-3 & 1.14 & 12.0 & 1.08 & 24.2 & 0.97 & 34.5 \\
& \textsc{EcoSpec} & \textbf{1.27} & \textbf{10.6} & \textbf{1.15} & \textbf{20.8} & \textbf{1.05} & \textbf{29.0} \\
\midrule
\multirow{2}{*}{AMC22-24} 
& EAGLE-3 & 1.14 & 11.9 & 1.07 & 20.7 & 1.03 & 31.3 \\
& \textsc{EcoSpec} & \textbf{1.45} & \textbf{10.9} & \textbf{1.30} & \textbf{18.6} & \textbf{1.09} & \textbf{27.9} \\
\midrule
\multirow{2}{*}{MTBench} 
& EAGLE-3 & 1.18 & 10.3 & 1.10 & 19.9 & 1.03 & 30.4 \\
& \textsc{EcoSpec} & \textbf{1.30} & \textbf{9.6} & \textbf{1.20} & \textbf{16.6} & \textbf{1.08} & \textbf{27.9} \\
\midrule
\multirow{2}{*}{Average}
& EAGLE-3 & 1.14 & 11.6 & 1.08 & 22.3 & 1.00 & 32.6 \\
& \textsc{EcoSpec} & \textbf{1.31} & \textbf{10.6} & \textbf{1.20} & \textbf{19.5} & \textbf{1.06} & \textbf{28.1} \\
\bottomrule
\end{tabular}
}
\end{table}

\subsection{Batch-Size Scaling of Verification Cost}
\label{sec:batch_size_verify_cost}

We further examine larger-batch behavior using a verification-cost-oriented metric on 100 randomly sampled instances from the seven evaluation datasets.
In Table~\ref{tab:batch_size_verify_cost}, we report
\[
\rho = \frac{T_{\mathrm{AR}}}{T_{\mathrm{verify}}},
\]
where $T_{\mathrm{AR}}$ denotes the average AR decoding time and $T_{\mathrm{verify}}$ denotes the average speculative verification time per step. A larger $\rho$ indicates that speculative verification is cheaper relative to AR decoding. We also report BS=8 Spd, the throughput of each speculative method normalized by AR throughput at batch size 8.

The ratio $\rho$ decreases for both methods as batch size increases.
On Qwen3-235B-A22B, $\rho$ drops from $0.59\times$ to $0.32\times$ for EAGLE-3 and from $0.67\times$ to $0.33\times$ for \textsc{EcoSpec}.
On GPT-OSS-120B, it drops from $0.72\times$ to $0.43\times$ for EAGLE-3 and from $0.90\times$ to $0.44\times$ for \textsc{EcoSpec}.
At BS=8, both speculative methods fall below AR throughput in this prototype, with BS=8 speedup of $0.69\times$ on Qwen3-235B-A22B and $0.65\times$ on GPT-OSS-120B.

These results indicate that large-batch speculative decoding remains challenging in this prototype.
Nevertheless, \textsc{EcoSpec} consistently maintains a higher $T_{\mathrm{AR}}/T_{\mathrm{verify}}$ ratio than EAGLE-3 across the tested batch sizes, indicating lower verification cost under the same setting.

\begin{table}[t]
\centering
\caption{Batch-size scaling of verification cost. We report $\rho=T_{\mathrm{AR}}/T_{\mathrm{verify}}$, the ratio between average AR decoding time and average speculative verification time per step. BS=8 Spd reports throughput normalized by AR throughput at batch size 8.}
\label{tab:batch_size_verify_cost}
\resizebox{\linewidth}{!}{
\begin{tabular}{llccccc}
\toprule
Model & Method & BS=1 & BS=2 & BS=4 & BS=8 & BS=8 Spd \\
\midrule
\multirow{2}{*}{Qwen3-235B-A22B}
& EAGLE-3 & 0.59$\times$ & 0.55$\times$ & 0.46$\times$ & 0.32$\times$ & 0.69$\times$ \\
& \textsc{EcoSpec} & \textbf{0.67$\times$} & \textbf{0.62$\times$} & \textbf{0.52$\times$} & \textbf{0.33$\times$} & 0.69$\times$ \\
\midrule
\multirow{2}{*}{GPT-OSS-120B}
& EAGLE-3 & 0.72$\times$ & 0.69$\times$ & 0.63$\times$ & 0.43$\times$ & 0.65$\times$ \\
& \textsc{EcoSpec} & \textbf{0.90$\times$} & \textbf{0.82$\times$} & \textbf{0.72$\times$} & \textbf{0.44$\times$} & 0.65$\times$ \\
\bottomrule
\end{tabular}
}
\end{table}

\section{Predictor Details and Analysis}

\subsection{Predictor Training Details}
\label{sec:predictor_training}

\textsc{EcoSpec} trains a lightweight expert predictor for each target MoE model to estimate expert activation cost before target-model verification.
The predictor is used only for cost-aware draft selection. It does not participate in token verification, does not query the target MoE model online, and does not modify the standard speculative verification rule.

\paragraph{Predictor backbones.}
We instantiate $\Pi_\theta$ with small open-source LMs and fine-tune them to predict the target model's per-layer expert activations from token-level inputs.
For DeepSeek-V3.1, we use DeepSeek-R1-Distill-Qwen-1.5B as the predictor backbone.
For Qwen3-235B-A22B, we use Qwen3-0.6B, leveraging architectural proximity within the Qwen family.
For GPT-OSS-120B, since no lightweight model from the same series is publicly available, we also use Qwen3-0.6B. The predictor cost is small relative to target-model verification, as reflected in Table~\ref{tab:latency}.

\paragraph{Training data and setup.}
We collect routing traces offline by running each target MoE model on seven datasets spanning reasoning, coding, and dialogue: HumanEval, MMStar, MT-Bench, AMC22-24, GSM8K, AIME-25, and Math500.
For each token and MoE layer, we record the target router's Top-$K$ selected experts as supervision, where $K$ follows the target model's routing configuration.
We randomly split samples into 80\% for training and 20\% for testing.
All evaluations reported in the main paper are conducted on the held-out test split.
We fine-tune predictors for 100 epochs using Adam with learning rate $1\times 10^{-5}$ and batch size 16.

\paragraph{Routing prediction accuracy.}
We report Top-$K$ routing prediction accuracy on the test split, computed as the average overlap ratio between the predicted Top-$K$ experts and the target router's Top-$K$ experts across tokens and MoE layers.
The resulting accuracies are 80\% for DeepSeek-V3.1, 82\% for Qwen3-235B-A22B, and 93\% for GPT-OSS-120B.
The higher accuracy on GPT-OSS-120B is consistent with its Top-4-over-128 routing setting, which yields more concentrated activation patterns than the Top-8 routing used by DeepSeek-V3.1 and Qwen3-235B-A22B.
Since EcoSpec uses the predictor only as a cost-estimation module, exact routing reconstruction is not required; the sensitivity to predictor quality is further evaluated in Appendix~\ref{sec:robustness}.

\subsection{Impact of Predictor Accuracy}
\label{sec:robustness}

To examine the effect of predictor quality, we evaluate \textsc{EcoSpec} with three predictor checkpoints: two intermediate checkpoints with Top-$K$ routing accuracy of approximately 50\% and 60\%, and the converged checkpoint used in the main experiments. The decoding configuration is fixed across all runs, including the default verification budget $\gamma=4$; only the predictor checkpoint is changed.

Table~\ref{tab:predictor_robustness} reports end-to-end speedup, acceptance length, and active expert count on Qwen3-235B-A22B and GPT-OSS-120B.
Lower-accuracy predictors lead to lower speedup and higher active expert counts.
On Qwen3-235B-A22B, the converged predictor improves the speedup from $1.17\times$ to $1.39\times$ on GSM8K and from $1.20\times$ to $1.62\times$ on MT-Bench compared with the 50\%-accuracy checkpoint.
On GPT-OSS-120B, the corresponding speedup improves from $1.01\times$ to $1.11\times$ on GSM8K and from $1.07\times$ to $1.30\times$ on MT-Bench.
The active expert count also decreases consistently as predictor quality improves.

These results show that predictor accuracy affects the quality of cost-aware draft selection.
The converged predictor gives the best speedup in all reported settings, while intermediate predictors still retain positive speedup under the same decoding configuration.

\begin{table}[t]
\centering
\caption{Robustness to Predictor Accuracy. Predictor-accuracy sensitivity. We vary the predictor checkpoint while keeping the decoding configuration fixed.}

\label{tab:predictor_robustness}
\resizebox{\linewidth}{!}{
\begin{tabular}{l|c|c|c|c}
\toprule
\multirow{2}{*}{\textbf{Model}} & \multirow{2}{*}{\textbf{Dataset}} & \textbf{Acc $\approx$ 50\%} & \textbf{Acc $\approx$ 60\%} & \cellcolor{gray!15}\textbf{Converged} \\
& & Spd / Len / Exp & Spd / Len / Exp & \textbf{Spd} / Len / Exp \\
\midrule
\multirow{2}{*}{\textbf{Qwen3-235B}} 
& GSM8K   & 1.17 / 2.20 / 24.5 & 1.29 / 2.50 / 22.5 & \cellcolor{gray!15}\textbf{1.39} / 2.54 / 21.0 \\
& MT-Bench& 1.20 / 2.45 / 25.2 & 1.45 / 2.50 / 23.8 & \cellcolor{gray!15}\textbf{1.62} / 2.44 / 19.9 \\
\midrule
\multirow{2}{*}{\textbf{GPT-OSS-120B}} 
& GSM8K   & 1.01 / 1.25 / 13.5 & 1.04 / 1.45 / 11.6 & \cellcolor{gray!15}\textbf{1.11} / 1.52 / 10.6 \\
& MT-Bench& 1.07 / 1.80 / 17.8 & 1.14 / 1.45 / 13.5 & \cellcolor{gray!15}\textbf{1.30} / 1.81 / 9.6 \\
\bottomrule
\end{tabular}
}
\end{table}

\subsection{Oracle Expert-Set Analysis}
\label{sec:oracle_predictor}

We further analyze whether the learned predictor provides a sufficiently accurate expert-cost signal for cost-aware draft selection.
In this analysis, we keep the \textsc{EcoSpec} selection rule unchanged and replace the predicted expert sets with ground-truth expert sets collected offline from the target MoE model.
This isolates the effect of predictor error while keeping the selection rule fixed.

The oracle expert sets are used only for analysis.
In normal speculative decoding, the true expert sets of candidate draft nodes are unavailable before target-model verification unless additional target-model computation is performed.
Therefore, this setting is reported with verification-oriented metrics rather than end-to-end throughput, and is not used as an online inference baseline.

\begin{table}[t]
\centering
\caption{Oracle expert-set analysis. Oracle uses ground-truth target-router activations collected offline, while Predictor uses the learned expert predictor. All entries report verification-oriented metrics under the same \textsc{EcoSpec} selection rule.}
\label{tab:oracle_predictor}
\resizebox{\linewidth}{!}{
\begin{tabular}{llccc}
\toprule
Model & Selector & Time (s) & Accept Len & Active Experts \\
\midrule
\multirow{2}{*}{Qwen3-235B-A22B}
& Oracle   & 0.730 & 2.32 & 20.4 \\
& Predictor & 0.730 & 2.32 & 20.5 \\
\midrule
\multirow{2}{*}{GPT-OSS-120B}
& Oracle   & 0.090 & 1.86 & 10.5 \\
& Predictor & 0.090 & 1.86 & 10.6 \\
\midrule
\multirow{2}{*}{DeepSeek-V3.1}
& Oracle   & 0.929 & 2.79 & 31.2 \\
& Predictor & 0.930 & 2.79 & 31.2 \\
\bottomrule
\end{tabular}
}
\end{table}

As shown in Table~\ref{tab:oracle_predictor}, replacing the learned predictor with oracle expert sets produces nearly unchanged acceptance lengths, verification times, and active expert counts.
This indicates that the converged predictor provides a cost signal close to that obtained from oracle expert sets in the evaluated settings.

This analysis complements the predictor-accuracy study.
The accuracy study shows that lower-quality predictors can reduce the effectiveness of cost-aware selection, while the oracle expert-set analysis shows that replacing the converged predictor with ground-truth expert sets yields little additional change.
Together, these results indicate that predictor quality matters, but the converged predictor is already sufficiently accurate for the selection rule used by \textsc{EcoSpec}.
\section{Evaluation under Different Verification Budgets ($\gamma$)}
\label{sec:sensitivity}

We vary the verification budget $\gamma \in \{3,4,5,6\}$ to support the common default setting used in the main experiments. Tables~\ref{tab:gamma_qwen} and~\ref{tab:gamma_gpt} report end-to-end speedup, acceptance length, and active expert count for EAGLE-3 and \textsc{EcoSpec} under each budget.

We select the default verification budget according to average end-to-end speedup across the seven datasets.
Under this criterion, $\gamma=4$ gives the highest average end-to-end speedup for both EAGLE-3 and \textsc{EcoSpec} on Qwen3-235B-A22B and GPT-OSS-120B. Increasing $\gamma$ beyond $4$ further increases acceptance length in many cases, but does not improve the averaged end-to-end speedup.

The sweep also shows that \textsc{EcoSpec} maintains its advantage over EAGLE-3 across the tested budgets.
On both target models, \textsc{EcoSpec} consistently achieves higher average end-to-end speedup than EAGLE-3 under $\gamma \in \{3,4,5,6\}$. It also uses fewer active experts on average under the same verification budget, showing that the benefit of cost-aware draft selection is not tied to a single budget choice.

Therefore, we use $\gamma=4$ as the default verification budget for both EAGLE-3 and \textsc{EcoSpec} in the main experiments.

\begin{table}[t]
\centering
\caption{Verification-budget sensitivity on Qwen3-235B-A22B. Each entry reports Spd / Len / Exp, corresponding to end-to-end speedup, acceptance length, and active expert count.}
\label{tab:gamma_qwen}
\resizebox{\linewidth}{!}{
\begin{tabular}{llcccc}
\toprule
\multirow{2}{*}{Dataset} 
& \multirow{2}{*}{Method} 
& $\gamma=3$ 
& \cellcolor{blue!5}$\gamma=4$ (Default) 
& $\gamma=5$ 
& $\gamma=6$ \\
& & Spd / Len / Exp & \cellcolor{blue!5}Spd / Len / Exp & Spd / Len / Exp & Spd / Len / Exp \\
\midrule
\multirow{2}{*}{GSM8K} 
& \cellcolor{gray!10}EAGLE-3 & \cellcolor{gray!10}1.25 / 2.45 / 21.8 & \cellcolor{gray!10}1.31 / 2.54 / 22.3 & \cellcolor{gray!10}1.15 / 2.62 / 27.5 & \cellcolor{gray!10}1.02 / 2.68 / 31.2 \\
& \textsc{EcoSpec} & 1.28 / 2.15 / 16.8 & \cellcolor{blue!5}\textbf{1.39} / 2.54 / 21.0 & 1.25 / 2.62 / 24.5 & 1.09 / 2.68 / 28.2 \\
\midrule
\multirow{2}{*}{HumanEval} 
& \cellcolor{gray!10}EAGLE-3 & \cellcolor{gray!10}1.08 / 2.25 / 20.5 & \cellcolor{gray!10}1.11 / 2.35 / 26.2 & \cellcolor{gray!10}1.02 / 2.41 / 30.8 & \cellcolor{gray!10}0.95 / 2.45 / 33.5 \\
& \textsc{EcoSpec} & 1.22 / 2.05 / 17.2 & \cellcolor{blue!5}\textbf{1.33} / 2.32 / 21.6 & 1.18 / 2.41 / 25.8 & 1.05 / 2.45 / 29.1 \\
\midrule
\multirow{2}{*}{MMStar} 
& \cellcolor{gray!10}EAGLE-3 & \cellcolor{gray!10}1.30 / 2.55 / 19.8 & \cellcolor{gray!10}1.35 / 2.64 / 20.5 & \cellcolor{gray!10}1.22 / 2.70 / 25.2 & \cellcolor{gray!10}1.08 / 2.75 / 29.5 \\
& \textsc{EcoSpec} & 1.41 / 2.25 / 15.1 & \cellcolor{blue!5}\textbf{1.57} / 2.58 / 18.4 & 1.38 / 2.65 / 22.5 & 1.15 / 2.70 / 26.2 \\
\midrule
\multirow{2}{*}{AIME-25} 
& \cellcolor{gray!10}EAGLE-3 & \cellcolor{gray!10}1.10 / 2.30 / 21.5 & \cellcolor{gray!10}1.14 / 2.41 / 22.1 & \cellcolor{gray!10}1.05 / 2.45 / 27.8 & \cellcolor{gray!10}0.96 / 2.50 / 30.5 \\
& \textsc{EcoSpec} & 1.15 / 2.08 / 16.5 & \cellcolor{blue!5}\textbf{1.23} / 2.37 / 20.9 & 1.12 / 2.44 / 24.2 & 1.02 / 2.48 / 27.8 \\
\midrule
\multirow{2}{*}{Math500} 
& \cellcolor{gray!10}EAGLE-3 & \cellcolor{gray!10}1.02 / 1.85 / 20.5 & \cellcolor{gray!10}1.05 / 1.96 / 25.4 & \cellcolor{gray!10}0.98 / 2.02 / 29.5 & \cellcolor{gray!10}0.90 / 2.05 / 32.8 \\
& \textsc{EcoSpec} & 1.08 / 1.75 / 15.9 & \cellcolor{blue!5}\textbf{1.13} / 1.88 / 19.4 & 1.05 / 1.95 / 23.8 & 0.98 / 2.01 / 27.5 \\
\midrule
\multirow{2}{*}{AMC22-24} 
& \cellcolor{gray!10}EAGLE-3 & \cellcolor{gray!10}1.08 / 2.10 / 21.2 & \cellcolor{gray!10}1.12 / 2.20 / 25.4 & \cellcolor{gray!10}1.04 / 2.25 / 30.1 & \cellcolor{gray!10}0.95 / 2.28 / 33.2 \\
& \textsc{EcoSpec} & 1.19 / 1.90 / 17.5 & \cellcolor{blue!5}\textbf{1.28} / 2.10 / 22.2 & 1.15 / 2.18 / 26.4 & 1.04 / 2.22 / 29.8 \\
\midrule
\multirow{2}{*}{MTBench} 
& \cellcolor{gray!10}EAGLE-3 & \cellcolor{gray!10}1.40 / 2.65 / 21.5 & \cellcolor{gray!10}1.46 / 2.78 / 24.0 & \cellcolor{gray!10}1.32 / 2.85 / 28.5 & \cellcolor{gray!10}1.15 / 2.90 / 32.5 \\
& \textsc{EcoSpec} & 1.48 / 2.40 / 16.2 & \cellcolor{blue!5}\textbf{1.62} / 2.44 / 19.9 & 1.45 / 2.55 / 24.1 & 1.20 / 2.60 / 28.5 \\
\midrule
\multirow{2}{*}{Average} 
& \cellcolor{gray!10}EAGLE-3 & \cellcolor{gray!10}1.18 / 2.31 / 21.0 & \cellcolor{gray!10}1.22 / 2.41 / 23.7 & \cellcolor{gray!10}1.11 / 2.47 / 28.5 & \cellcolor{gray!10}1.00 / 2.52 / 31.9 \\
& \textsc{EcoSpec} & 1.26 / 2.08 / 16.5 & \cellcolor{blue!5}\textbf{1.36} / 2.32 / 20.5 & 1.22 / 2.40 / 24.5 & 1.08 / 2.45 / 28.2 \\
\bottomrule
\end{tabular}
}
\end{table}

\begin{table}[t]
\centering
\caption{Verification-budget sensitivity on GPT-OSS-120B.}
\label{tab:gamma_gpt}
\resizebox{\linewidth}{!}{
\begin{tabular}{llcccc}
\toprule
\multirow{2}{*}{Dataset} 
& \multirow{2}{*}{Method} 
& $\gamma=3$ 
& \cellcolor{blue!5}$\gamma=4$ (Default) 
& $\gamma=5$ 
& $\gamma=6$ \\
& & Spd / Len / Exp & \cellcolor{blue!5}Spd / Len / Exp & Spd / Len / Exp & Spd / Len / Exp \\
\midrule
\multirow{2}{*}{GSM8K} 
& \cellcolor{gray!10}EAGLE-3 & \cellcolor{gray!10}1.02 / 1.45 / 11.2 & \cellcolor{gray!10}1.05 / 1.56 / 11.9 & \cellcolor{gray!10}1.01 / 1.62 / 13.5 & \cellcolor{gray!10}0.96 / 1.65 / 15.2 \\
& \textsc{EcoSpec} & 1.20 / 1.30 / 8.5 & \cellcolor{blue!5}1.11 / 1.52 / 10.6 & 1.08 / 1.60 / 11.5 & 1.02 / 1.65 / 13.2 \\
\midrule
\multirow{2}{*}{HumanEval} 
& \cellcolor{gray!10}EAGLE-3 & \cellcolor{gray!10}1.02 / 1.62 / 11.0 & \cellcolor{gray!10}1.05 / 1.72 / 11.8 & \cellcolor{gray!10}1.00 / 1.78 / 13.2 & \cellcolor{gray!10}0.94 / 1.82 / 15.0 \\
& \textsc{EcoSpec} & 1.15 / 1.45 / 8.8 & \cellcolor{blue!5}\textbf{1.26} / 1.72 / 10.7 & 1.18 / 1.78 / 11.9 & 1.09 / 1.82 / 13.5 \\
\midrule
\multirow{2}{*}{MMStar} 
& \cellcolor{gray!10}EAGLE-3 & \cellcolor{gray!10}1.22 / 1.80 / 11.1 & \cellcolor{gray!10}1.28 / 1.91 / 11.7 & \cellcolor{gray!10}1.15 / 1.98 / 13.5 & \cellcolor{gray!10}1.05 / 2.05 / 15.5 \\
& \textsc{EcoSpec} & 1.32 / 1.62 / 8.2 & \cellcolor{blue!5}\textbf{1.45} / 1.90 / 10.5 & 1.30 / 1.98 / 11.8 & 1.15 / 2.05 / 13.8 \\
\midrule
\multirow{2}{*}{AIME-25} 
& \cellcolor{gray!10}EAGLE-3 & \cellcolor{gray!10}1.12 / 1.92 / 11.5 & \cellcolor{gray!10}1.17 / 2.02 / 12.2 & \cellcolor{gray!10}1.08 / 2.10 / 13.8 & \cellcolor{gray!10}0.98 / 2.15 / 15.9 \\
& \textsc{EcoSpec} & 1.21 / 1.75 / 8.6 & \cellcolor{blue!5}\textbf{1.33} / 2.02 / 10.8 & 1.21 / 2.10 / 12.2 & 1.08 / 2.15 / 14.1 \\
\midrule
\multirow{2}{*}{Math500} 
& \cellcolor{gray!10}EAGLE-3 & \cellcolor{gray!10}1.10 / 1.92 / 11.2 & \cellcolor{gray!10}1.14 / 2.02 / 12.0 & \cellcolor{gray!10}1.05 / 2.08 / 13.6 & \cellcolor{gray!10}0.95 / 2.12 / 15.8 \\
& \textsc{EcoSpec} & 1.18 / 1.72 / 8.4 & \cellcolor{blue!5}\textbf{1.27} / 2.00 / 10.6 & 1.15 / 2.08 / 11.9 & 1.05 / 2.12 / 14.0 \\
\midrule
\multirow{2}{*}{AMC22-24} 
& \cellcolor{gray!10}EAGLE-3 & \cellcolor{gray!10}1.10 / 1.92 / 11.1 & \cellcolor{gray!10}1.14 / 2.02 / 11.9 & \cellcolor{gray!10}1.04 / 2.10 / 13.8 & \cellcolor{gray!10}0.96 / 2.15 / 15.7 \\
& \textsc{EcoSpec} & 1.32 / 1.75 / 8.9 & \cellcolor{blue!5}\textbf{1.45} / 2.02 / 10.9 & 1.30 / 2.10 / 12.4 & 1.09 / 2.15 / 14.5 \\
\midrule
\multirow{2}{*}{MTBench} 
& \cellcolor{gray!10}EAGLE-3 & \cellcolor{gray!10}1.12 / 1.82 / 9.2 & \cellcolor{gray!10}1.18 / 1.91 / 10.3 & \cellcolor{gray!10}1.08 / 1.96 / 11.5 & \cellcolor{gray!10}0.98 / 2.02 / 13.2 \\
& \textsc{EcoSpec} & 1.20 / 1.55 / 8.1 & \cellcolor{blue!5}\textbf{1.30} / 1.81 / 9.6 & 1.20 / 1.88 / 11.2 & 1.08 / 1.92 / 12.8 \\
\midrule
\multirow{2}{*}{Average} 
& \cellcolor{gray!10}EAGLE-3 & \cellcolor{gray!10}1.10 / 1.72 / 10.9 & \cellcolor{gray!10}1.14 / 1.88 / 11.6 & \cellcolor{gray!10}1.06 / 1.96 / 13.3 & \cellcolor{gray!10}0.97 / 2.02 / 15.1 \\
& \textsc{EcoSpec} & 1.23 / 1.59 / 8.5 & \cellcolor{blue!5}\textbf{1.31} / 1.86 / 10.6 & 1.20 / 1.93 / 11.8 & 1.08 / 1.98 / 13.7 \\ 
\bottomrule
\end{tabular}
}
\end{table}

\section{Expert Activation Pattern Analysis}
\label{sec:load_balance}

\begin{figure*}[t]
    \centering
    \begin{subfigure}[b]{0.48\textwidth}
        \centering
        \includegraphics[width=\textwidth]{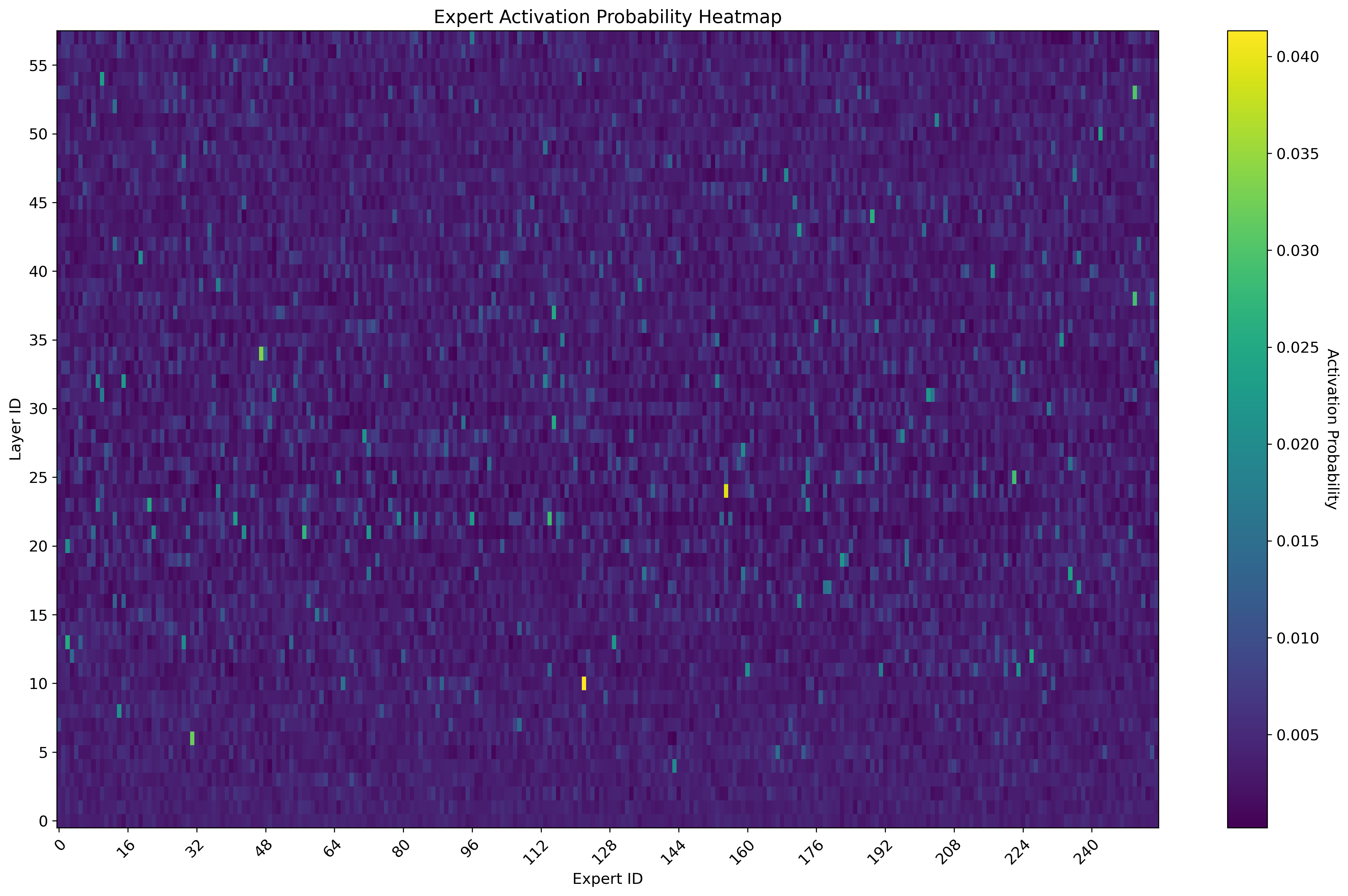}
        \caption{\textbf{DeepSeek-V3.1}: Activation Heatmap}
        \label{fig:ds_heat}
    \end{subfigure}
    \hfill
    \begin{subfigure}[b]{0.48\textwidth}
        \centering
        \includegraphics[width=\textwidth]{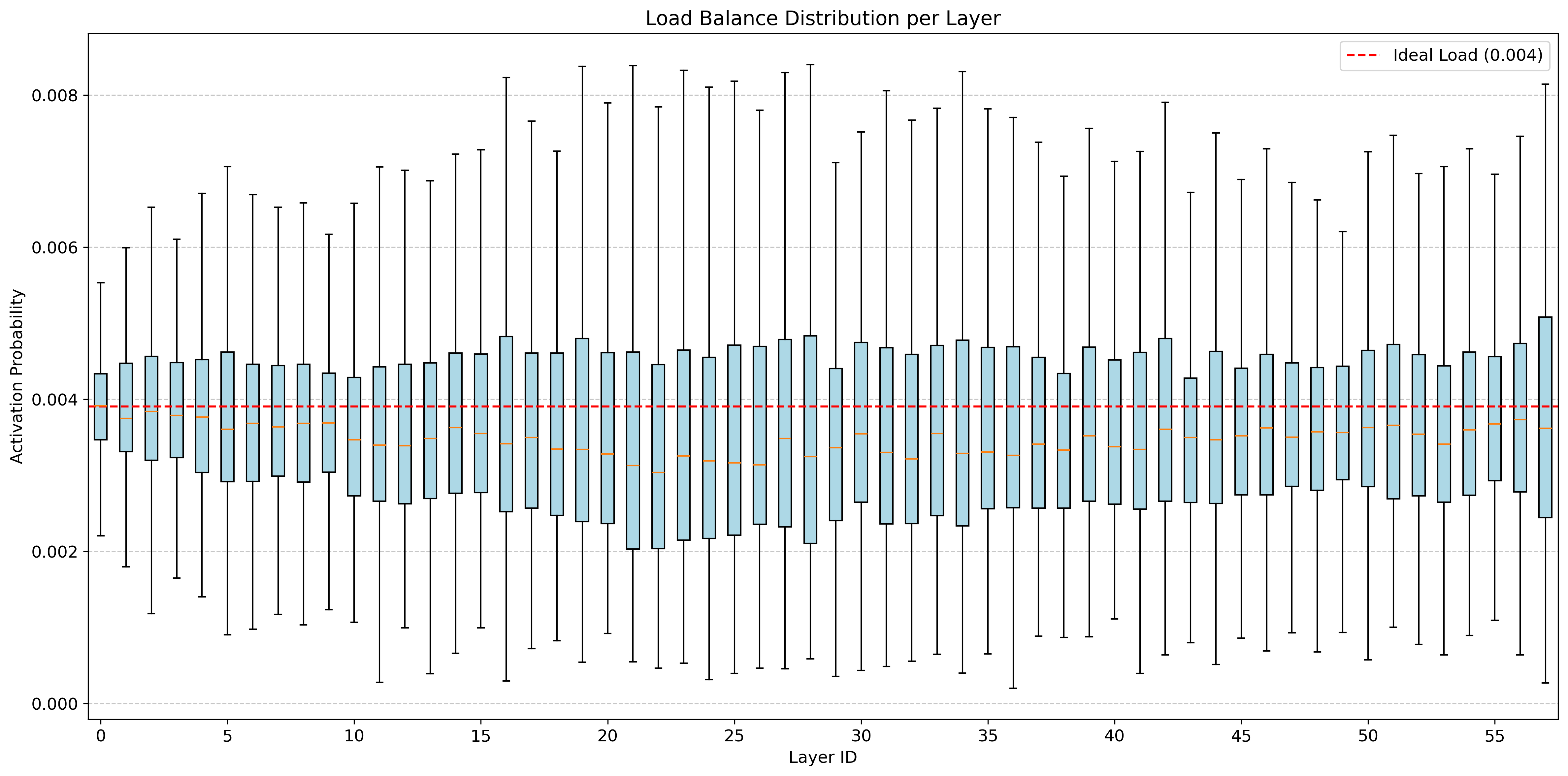}
        \caption{\textbf{DeepSeek-V3.1}: Load Distribution}
        \label{fig:ds_dist}
    \end{subfigure}
    
    \begin{subfigure}[b]{0.48\textwidth}
        \centering
        \includegraphics[width=\textwidth]{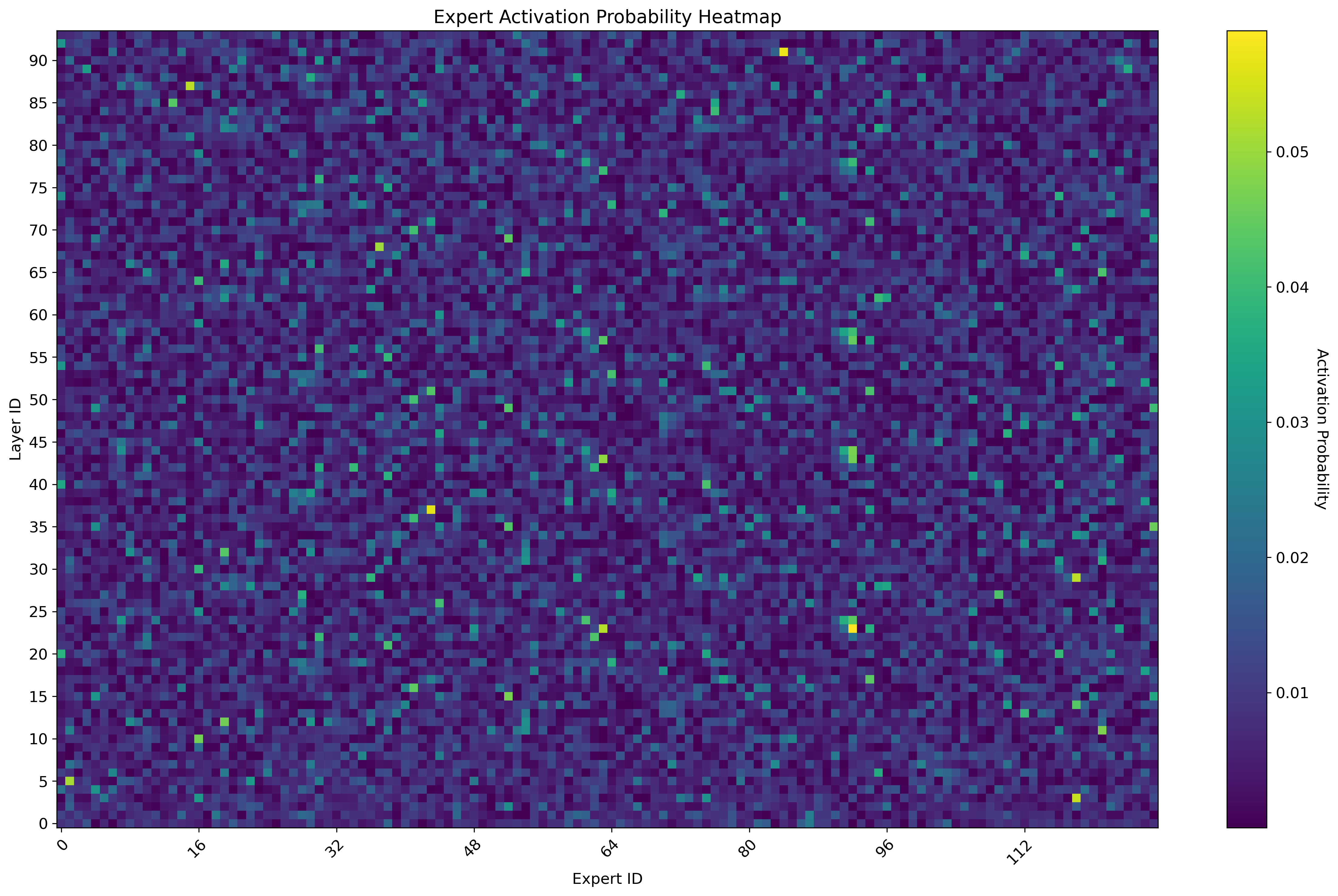}
        \caption{\textbf{Qwen3-235B}: Activation Heatmap}
        \label{fig:qwen_heat}
    \end{subfigure}
    \hfill
    \begin{subfigure}[b]{0.48\textwidth}
        \centering
        \includegraphics[width=\textwidth]{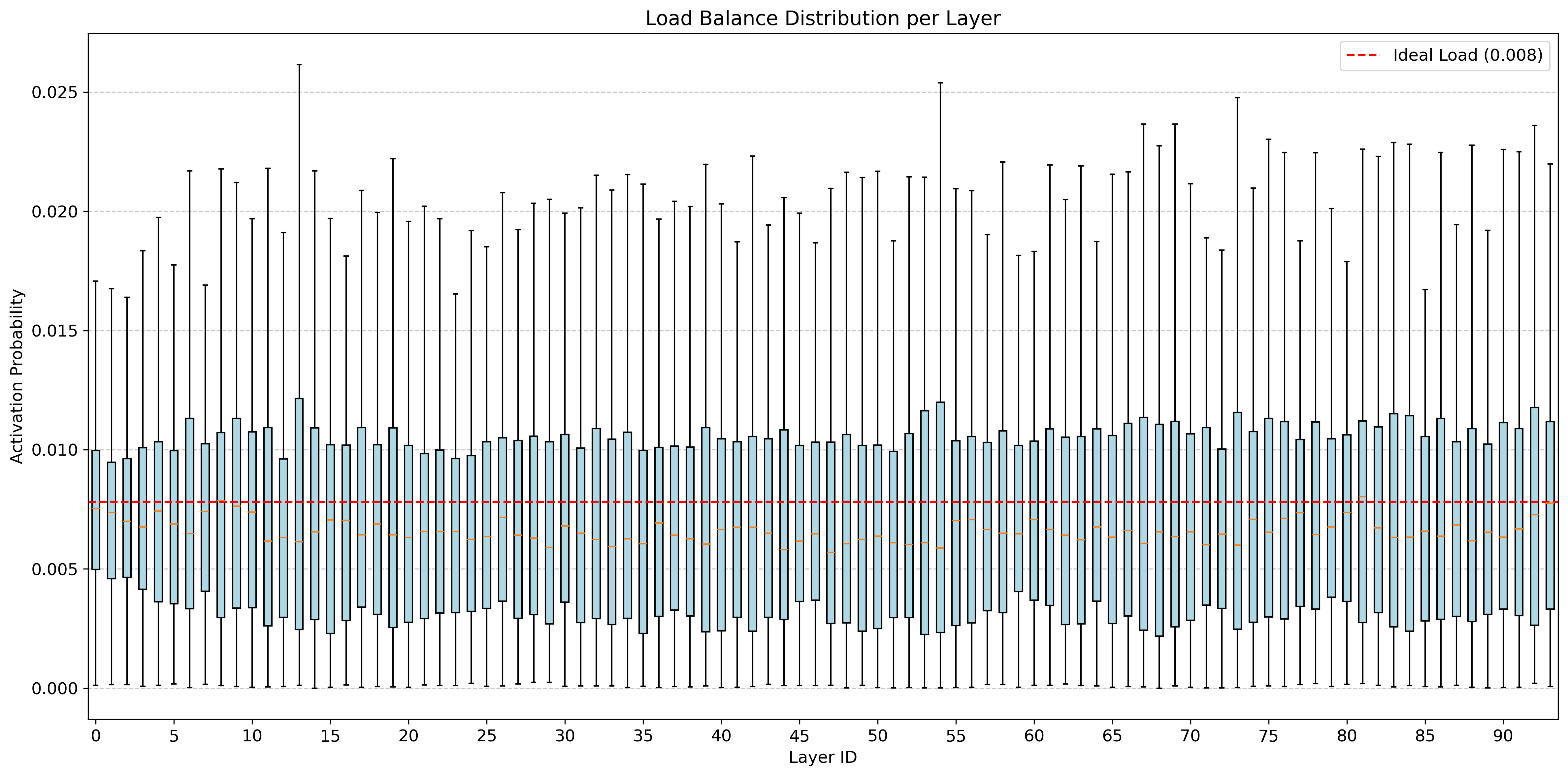}
        \caption{\textbf{Qwen3-235B}: Load Distribution}
        \label{fig:qwen_dist}
    \end{subfigure}
    
    \begin{subfigure}[b]{0.48\textwidth}
        \centering
        \includegraphics[width=\textwidth]{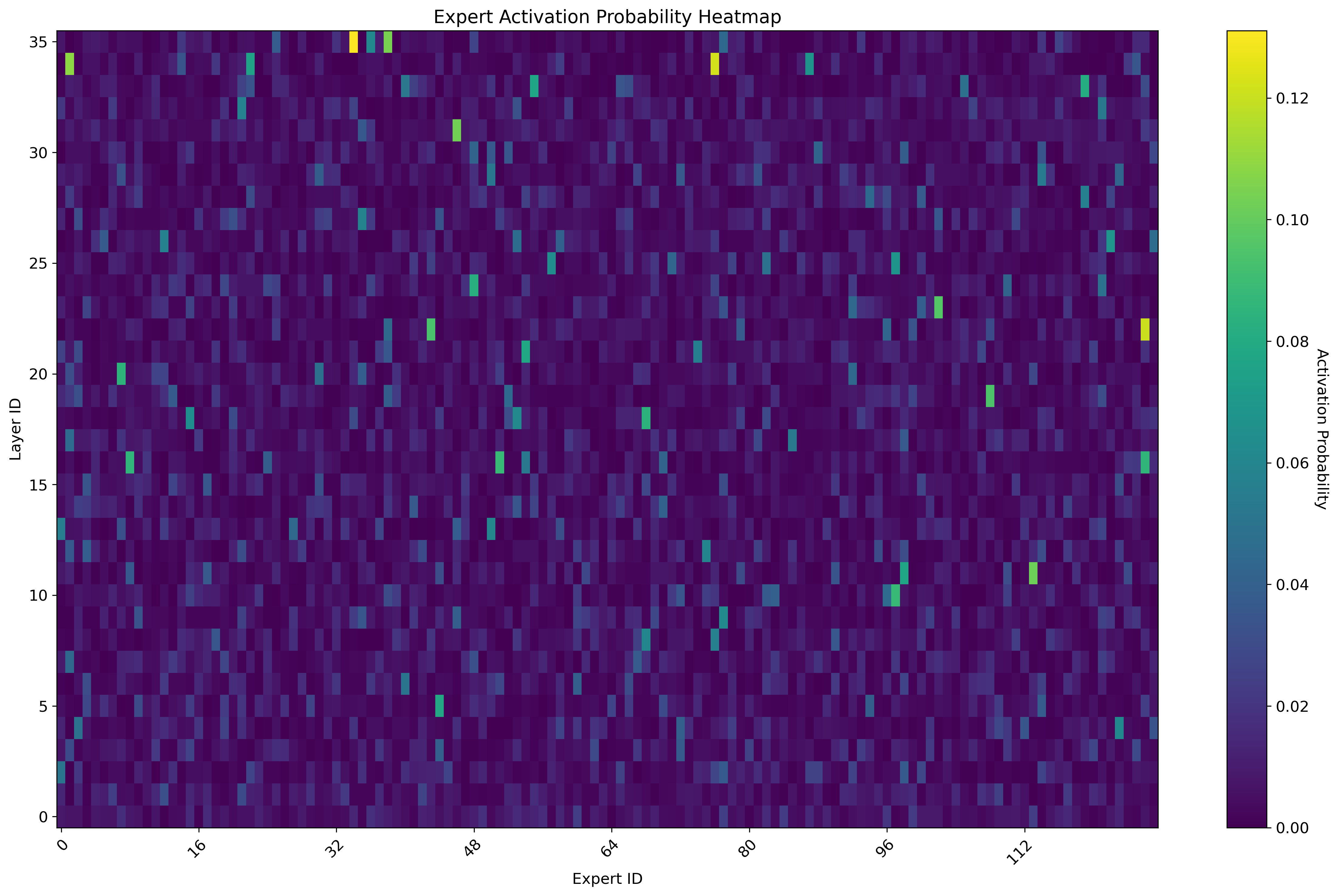}
        \caption{\textbf{GPT-OSS-120B}: Activation Heatmap}
        \label{fig:gpt_heat}
    \end{subfigure}
    \hfill
    \begin{subfigure}[b]{0.48\textwidth}
        \centering
        \includegraphics[width=\textwidth]{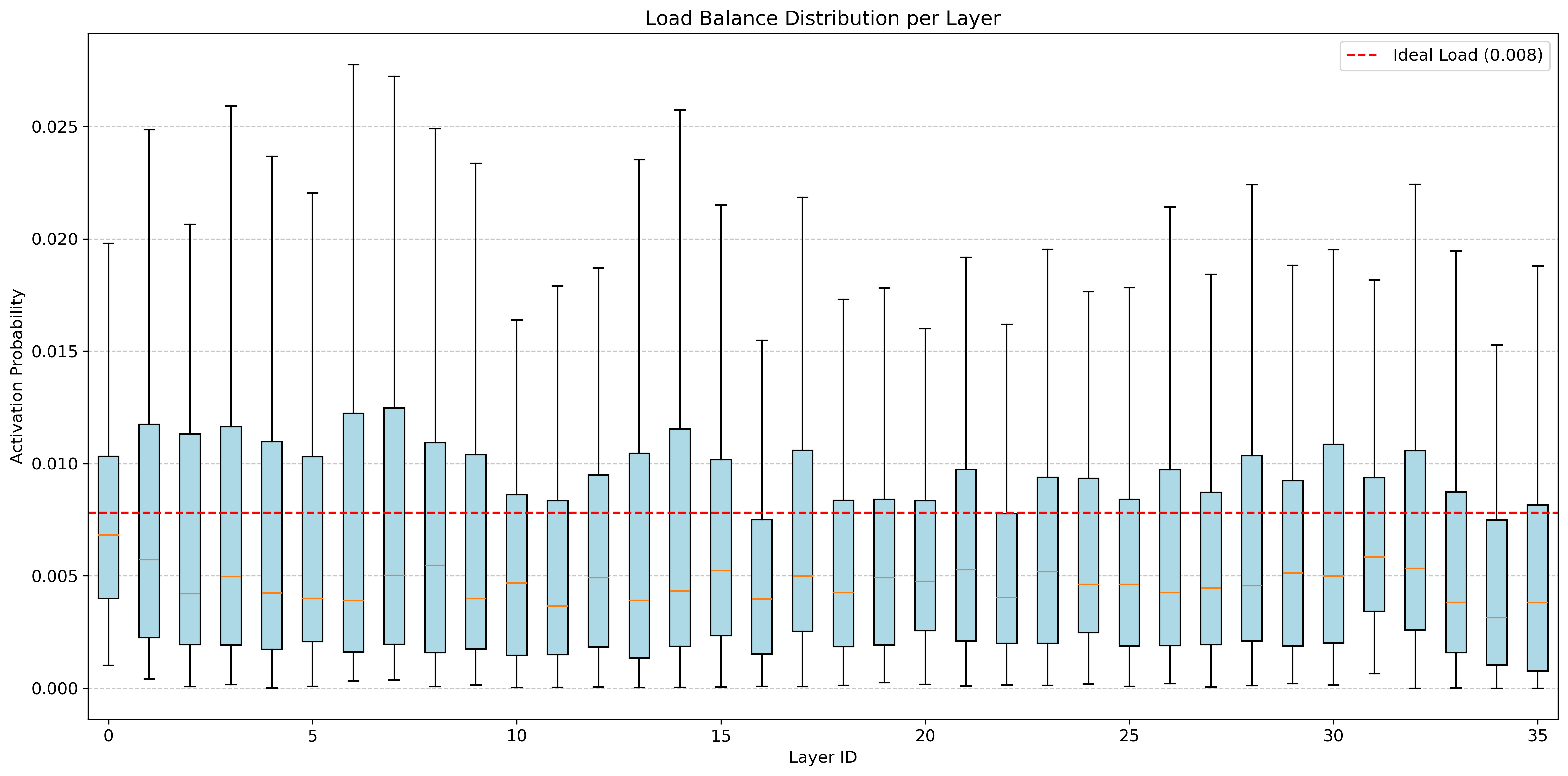}
        \caption{\textbf{GPT-OSS-120B}: Load Distribution}
        \label{fig:gpt_dist}
    \end{subfigure}
    \caption{Expert activation patterns across evaluated MoE models. Each row corresponds to one target model. The left panel shows the expert-activation heatmap, and the right panel shows the layer-wise expert-load distribution.}
    \label{fig:expert_analysis_combined}
\end{figure*}

We analyze expert-activation patterns of the evaluated MoE backbones in Fig.~\ref{fig:expert_analysis_combined}.
For each model, the left panel shows the expert-activation heatmap across layers, and the right panel shows the layer-wise expert-load distribution.
This analysis characterizes the expert working sets encountered during speculative verification and provides context for why active expert count is reported alongside acceptance length.

\paragraph{DeepSeek-V3.1.}
DeepSeek-V3.1 shows a relatively diffuse activation pattern.
The heatmap does not concentrate on a small subset of experts, and the layer-wise load distributions are relatively narrow across most layers.
This indicates that expert usage is broadly balanced across routed experts.
For speculative verification, such balanced routing still makes the expert working set an important cost component: verifying multiple draft tokens can involve many distinct experts even when no small group of experts dominates the routing pattern.

\paragraph{Qwen3-235B-A22B and GPT-OSS-120B.}
Qwen3-235B-A22B and GPT-OSS-120B show more concentrated expert-activation patterns.
Their heatmaps contain clearer high-usage regions, and their layer-wise load distributions have wider ranges with more visible high-load experts.
These patterns indicate stronger locality in expert usage: different candidate tokens are more likely to share parts of their expert sets, making the verified expert working set sensitive to which draft tokens are selected.

\paragraph{Implication for cost-aware draft selection.}
These observations support treating the expert working set as an explicit cost component in MoE speculative decoding.
Acceptance length measures how many drafted tokens are verified successfully, but it does not describe which experts are activated during verification.
Two draft sets with similar acceptance length can induce different expert working sets and therefore different verification costs.
\textsc{EcoSpec} incorporates this cost dimension during draft selection by favoring candidates with lower predicted marginal expert cost under the same verification budget.

\section{Datasets and Evaluation Details}
\label{sec:dataset_details}

We evaluate \textsc{EcoSpec} on seven benchmarks covering mathematical reasoning, code generation, dialogue, and text-only inputs derived from a vision-language benchmark. All methods are evaluated with the same prompts and decoding settings on each dataset, so the reported speedup, acceptance length, and active expert count are computed under matched input conditions.

\paragraph{GSM8K}
GSM8K~\cite{gsm8k} contains grade-school math word problems that require multi-step arithmetic reasoning.
We use the standard 5-shot Chain-of-Thought setting, where five exemplars are prepended to the input before the test question.

\paragraph{AIME 2025} 
AIME 2025~\cite{aime2025} contains competition-level mathematical problems from the American Invitational Mathematics Examination. We use a zero-shot reasoning prompt and require the final answer to be placed in a boxed format:
\texttt{\{question\}\textbackslash n Please reason step by step, and put your final answer within \textbackslash boxed\{\}.}

\paragraph{MATH500 and AMC22-24}
MATH500~\cite{math500-1, math500-2} is a 500-problem subset of the MATH benchmark. AMC22-24~\cite{amc22_24} contains problems from the American Mathematics Competitions from 2022 to 2024.
For both datasets, we use the following zero-shot prompt:
\texttt{Problem:\textbackslash n \{problem\}\textbackslash n\textbackslash n Solution:}

\paragraph{HumanEval} 
HumanEval~\cite{humaneval} contains 164 Python programming problems with function signatures, docstrings, and unit tests.
We use the benchmark prompts as code-generation inputs.

\paragraph{MT-Bench}
MT-Bench~\cite{mtbench} is a multi-turn instruction-following benchmark.

\paragraph{MMStar} 
MMStar~\cite{mmstar} is a vision-language benchmark whose original evaluation involves both image and text inputs.
Since this work focuses on text-only LLM inference, we remove the image inputs and provide only the textual questions to the model.
This setting is used as a text-only generation workload and is not intended to measure vision-language grounding ability.

\section{Model Details}
\label{app:model_details}

Table~\ref{tab:model_summary} summarizes the MoE configurations of the target models used in our experiments.
We include the total and active parameter counts, the number of MoE layers, the number of experts per MoE layer, and the routing Top-$k$.
These attributes determine the scale of expert activation during verification and are directly related to the cost-aware draft selection studied in this paper.

\begin{table}[th]
\centering
\caption{Architectural summary of evaluated MoE models. Total and active parameters follow the reporting convention of the corresponding model releases. Top-$k$ denotes the number of selected experts per token.}
\label{tab:model_summary}
\resizebox{\linewidth}{!}{
\begin{tabular}{lccc}
\toprule
Model & DeepSeek-V3.1 & Qwen3-235B-A22B & GPT-OSS-120B \\
\midrule
Total / active parameters & 671B / 37B & 235B / 22B & 120B / 5.1B \\
Transformer blocks & 61 & 94 & 36 \\
MoE / dense layers & 58 / 3 & 94 / 0 & 36 / 0 \\
Experts per MoE layer & 256 routed + 1 shared & 128 & 128 \\
Top-$k$ experts per token & 8 & 8 & 4 \\
Maximum context length & 128K & 40,960 & 131,072 \\
\bottomrule
\end{tabular}
}
\end{table}


\end{document}